\documentclass[11pt]{article}

\usepackage{epsfig,amsmath,latexsym,amssymb}
\usepackage{graphicx}
\usepackage{lscape}
\usepackage{picture, eso-pic, tikz} 
\usepackage{dsfont}
\usepackage{appendix}
\usepackage{multirow}
\usepackage{rotating}
\usepackage{listings}
\usepackage{adjustbox}
\usepackage{etoolbox}
\BeforeBeginEnvironment{appendices}{\clearpage}

\renewcommand{\baselinestretch}{1.1}
\def\R{{\mathbb R}}  
\def\N{{\mathbb N}}  

\DeclareMathOperator*{\argminA}{arg\,min}

\newcommand{\Remm}[1]{}

\newtheorem{model ass}[theo]{Model Assumptions}

\numberwithin{equation}{section}

\definecolor{MyGray}{rgb}{0.92,0.92,0.92}


\def\bx{\boldsymbol{x}}

\def\bz{\boldsymbol{z}}
\def\bw{\boldsymbol{w}}

\def\bff{\boldsymbol{f}}

\begin{document}

	\author{	Salvatore Scognamiglio \footnote{Department of Management and Quantitative Sciences, University of Naples ``Parthenope",\newline salvatore.scognamiglio@uniparthenope.it}}

\date{Version of May 27, 2021}
\title{Calibrating the Lee-Carter and the Poisson Lee-Carter models  via Neural Networks}
\maketitle

\begin{abstract}
This paper introduces a neural network approach for fitting the Lee-Carter and the Poisson Lee-Carter model on multiple populations. 
We develop some neural networks that replicate the structure of the individual LC models and allow their joint fitting  by  analysing the mortality data of all the considered populations  simultaneously. 
The neural network architecture is specifically designed to calibrate each individual model  using all available information instead of using a population-specific subset of data as in the traditional estimation schemes. 
A large set of  numerical experiments performed on all the countries of the Human Mortality Database (HMD) shows the effectiveness of our approach. In particular,  the resulting parameter estimates appear smooth and less sensitive to the random fluctuations often present in the mortality rates' data, especially for low-population countries.  In addition, the forecasting performance results significantly improved as well. 

\end{abstract}

	{\bf Keywords.} Mortality modeling, Multi-population mortality modeling, neural networks, Lee--Carter model, Human Mortality Database.



\section{Introduction}
\label{sec:introduction}
In recent decades, the mortality of most developed countries was gradually declining as result of improvements in public health, medical advances, lifestyle changes and government regulation.
Despite it is an obvious benefit for society,  this longevity improvement could also represent a risk for governments and insurance companies.
Indeed, if they do not properly consider these improvements in retirement planning and the  life insurance products' pricing, they could get in financial trouble.    
The risk that future mortality and life expectancy outcomes turn out different than expected is typically called \emph{longevity risk} and, 
 as pointed out in \cite{Barrieu2012}, its management requires stochastic mortality projection models. 
In this vein, a number of stochastic mortality models  were developed. 

One of the first  stochastic models describing the mortality of a single population was proposed by Lee and Carter (LC) \cite{Lee1992}. 
Their model decomposes  the age-time matrix of mortality rates into a bilinear combination of age and period parameters using the Principal Component Analysis (PCA) and forecasting  is performed by projecting the time-index component into the future with time-series models. 
A formal description of this model will be presented in the next section. 

Numerous extensions of the LC model have been developed and proposed in literature.
For example, \cite{Brouhns2002a} embedded the LC model into a Poisson regression setting to overcome the homoskedastic error structure assumed into the original LC method. \cite{Renshaw2003} proposed a multi-factor version of the LC model to improve the goodness-of-fit and a few years later \cite{Renshaw2006} generalised the Lee-Carter model  including a cohort effect.
\cite{Hyndman2007} proposed a functional data approach in which the mortality curves are smoothed for each year using constrained regression splines prior to fitting a model using principal components decomposition and \cite{Hainaut2020} further extended this method by using a wavelet-based decomposition. 

Another very popular stochastic mortality model is the Cairns-Blake-Dowd (CBD) model proposed in \cite{Cairns2006}  and  many its extensions have been proposed. We refer to \cite{Cairns2009}  for a review.


Since the most of the drivers of the mortality improvements mentioned above often spread quickly, mortality changes over time between different countries appear, in some way, correlated.
For this reason, the study of multi-population mortality models has received increasing attention within the mortality forecasting'  literature. 
An extensive use of these models is typical in  reinsurance and risk hedging  \cite{Enchev2017, Villegas2017}. 
One of the simplest approaches for forecasting  mortality of multiple populations consists of  using Individual Lee-Carter (ILC) models \cite{Li2011}. 
In this case, the mortality of each population is described by an own LC model whose parameters are estimated separately from the other populations.
This approach is relatively accurate and easy-to-implement even for a large number of populations; however, it completely ignores the dependency among mortality  of the different populations. 
Some authors address this issue by introducing common terms in the individual  models. 
A very popular model is  the Augmented Common Factor (ACF) model developed by  Li and Lee \cite{Li2005} that proposes a  double log-bilinear mortality model augmenting common age and period effects with sub–population specific age and period effects. 
A second example is the Common Age Effect (CAE) model proposed by Kleinow \cite{Kleinow2015}, in which the component of the LC model describing the change in mortality with time is held constant, but different period indices are fit for each population. 
However, these models were usually intended for forecasting the mortality of similar populations and not for large-scale mortality forecasting. 
The comparative analysis in \cite{Richman2020} highlights that these multi-population extensions did not perform fully competitively against the ILC approach when a large set of populations is considered. 

Large-scale mortality forecasting  is defined as the simultaneous production of forecasts for many distinct populations and potentially different among them. 
Examples of large-scale mortality forecasting tasks are the forecasting the mortality rates of all the populations of the Human Mortality Database  (HMD) or the United States Mortality Database (USMD) simultaneously. Thanks to their ability to analyse large amounts of data and  model non-linearities,  neural networks represent a natural candidate to address this challenge. 
Although   neural networks' application to mortality modeling is quite recent, the  scientific contributions are increasing in number and intensity. 
\cite{Hainaut2018} proposed a neural network approach to predict and simulate mortality rates of a single population. The author developed a neural analyser for extracting latent time processes and directly predicting mortality. This approach allows for detecting and replicating non-linearities observed in the evolution of log-forces of mortality. 
The same intuition motivated the contribution in \cite{Nigri2019}, in which the authors introduced Recurrent Neural Networks (RNNs) into the  classical two-stage procedure of the LC approach.
In particular, they employed Long Short-Term Memory (LSTM) networks to model the time-related index component. Furthermore, \cite{Lindholm2021} explored the application of the LSTM networks  in the Poisson LC model framework.  
\cite{Richman2020}  has the merit of  developing   the first large-scale mortality model based on neural networks.
They provided a neural network architecture  based on fully-connected and embedding layers with notable forecasting accuracy.
\cite{Perla2021} further extended the model of \cite{Richman2020} by introducing RNNs and Convolutional Neural Networks (CNNs), specifically designed to model sequential data such as time-series data.  The use of convolutional networks for mortality modeling was also investigated in \cite{Wang2021}.
Despite the models proposed in \cite{Richman2020} and \cite{Perla2021} present more accurate forecasts than the ILC approach, how forecast uncertainty can be derived remains an open issue. 

This paper proposes a different approach to perform large-scale mortality forecasting. We use neural networks  for fitting  some well-known mortality models without modifying their forecasting scheme; this allows us to derive interval estimates. 
The main idea consists of developing neural network architectures that, on one side, replicate the model structure of the single-population mortality models and, at the same time, take into account and exploit the dependency among the mortality  of  different populations.
To this purpose, we embed the individual LC models into a neural network in which the classical LC parameters are jointly estimated  by processing  the mortality data of all populations simultaneously. 
The neural network architecture is specifically designed to calibrate each individual LC  using all available information instead of using a population-specific subset of data as in the traditional estimation schemes. 
We believe that this approach could produce estimates less sensitive to the random fluctuations often present in mortality rates' data. 
The neural network architectures developed present very few parameters to optimise and are easy-to-interpret. 
These features could  encourage the use of neural networks in mortality modeling also by practitioners who are wary  of the use of complex and hard-to-interpret models even if they have  high predictive power.
Despite the simple structure of the neural networks proposed, the forecasting performance is highly competitive with respect to other neural network-based approaches proposed in the literature. 

The remainder of the paper is structured as follows: Section \ref{sec:LCmodel} provides a formal description of the LC model, Section \ref{sec:NN} introduces the neural network architectures employed in this paper, Section \ref{sec:core} formally presents the  neural network-based model, in Section \ref{sec:numerical_experiments} a large set of numerical experiments is illustrated  and finally, Section \ref{sec:conclusions} concludes. 

\section{Lee-Carter Model}
\label{sec:LCmodel}
 The Lee-Carter (LC) model \cite{Lee1992} is an elegant and powerful approach to  forecast  a single population's mortality. 
Let $\mathcal{X} = \{x_{0}, x_{1}, \dots, x_{\omega}\}$ be the set of the age categories and  $\mathcal{T} = \{t_{0}, t_{1}, \dots, t_{n}\}$ be the set of calendar years considered. 
The LC model defines the logarithm of the central death rate  $log(m_{x,t}) \in \mathbb{R}$ at age $x \in \mathcal{X}$ in the calendar year $t\in \mathcal{T}$  as
\begin{equation}
    \label{eq:LC}
    log(m_{x,t}) = a_x+b_x k_t+ e_{x,t},
\end{equation}
where
 $a_x \in \mathbb{R}$ is the average age-specific pattern of mortality,
 $b_x\in \mathbb{R}$ represents the age-specific patterns of mortality change and indicates the sensitivity of the logarithm of the force of mortality at age $x$ to variations in the time index $k_t$,
$k_t\in \mathbb{R}$ is the time index describing mortality trend and $e_{x,t} \in \mathbb{R}$ is the error term.

Since the model  in \eqref{eq:LC} is over-parameterised, to  avoid identifiability problems,
  the following constraints are imposed 
\begin{equation}
\label{eq:constraints}
    \sum_{x \in \mathcal{X}} b_x = 1 \quad \quad \quad \sum_{t \in \mathcal{T}} \frac{k_t}{\mid \mathcal{T} \mid} = 0. 
\end{equation}
The Ordinary Least Squared (OLS) estimation of the model parameters in (\ref{eq:LC}) can be obtained by solving the optimisation problem 
\begin{equation}
\argminA_{(a_x)_x,(b_x)_x, (k_t)_t}  \sum_{x \in \mathcal{X}} \sum_{t \in \mathcal{T}} \bigg(log (m_{x,t})- a_x - b_x k_t \bigg)^2 \\.
\end{equation}

The $(a_x)_x$ are estimated as the logarithm of the geometric mean of the crude
mortality rates, averaged over all $t$, for each $x \in \mathcal{X}$
\begin{equation*}
    \hat{a}_x = log \bigg( \prod_{t\in \mathcal{T}} (m_{x,t})^{1/\mid \mathcal{T} \mid} \bigg),
\end{equation*}
while $(k_t)_t$  and $(b_x)_x$ are estimated as the first right and first left singular vectors in the SVD of the center log-mortality matrix $ M =     \big(log(m_{x,t}) - \hat{a}_x \big)_{ x \in \mathcal{X}, t \in \mathcal{T}}  \in \mathbb{R}^{|\mathcal{X}| \times |\mathcal{T}|}$.
In order to forecast, the parameters $(a_x)_x$ and $(b_x)_x$ are assumed to be constant over time while the time index $k_t$ is modeled as an ARIMA (0,1,0)  process 
\begin{equation}
k_t = k_{t-1}+ \gamma+ e_t \quad \quad with \ i.i.d \ e_t \sim N(0, \sigma_{\epsilon}^2) 
\label{eq:kt_arima}
\end{equation}
where $\gamma \in \mathbb{R}$ is the  drift.

A simple way of modeling the  mortality of a set of different populations $\mathcal{I}$  is to describe each population separately with its own LC model
\begin{equation}
    \label{eq:ILC}
    log(m^{(i)}_{x,t}) = a^{(i)}_x+b^{(i)}_x k^{(i)}_t+ e^{(i)}_{x,t} \quad \quad \forall i \in \mathcal{I}. 
\end{equation}
This approach is sometimes called  Individual Lee Carter (ILC) approach.  In this case,
the model fitting is performed individually  $\forall i \in \mathcal{I}$, 
and the population and time-specific terms $k_t^{(i)}$  are projected with independent ARIMA (0,1,0) processes.
Despite some multi-population extensions of the LC model were proposed, some comparative studies have highlighted that the ILC approach remains highly competitive and represents a reliable benchmark.

\subsection{Poisson Maximum Likelihood Estimation}
\label{sec:poisson_lc}
The main drawback of SVD is the assumption of homoskedastic errors \cite{Alho2000}.   
This issue is related to the fact that, for inference, we are actually assuming that the errors are normally distributed, which is quite unrealistic.  Indeed, 
appear reasonable to believe that the logarithm of the observed log-mortality rates is much more variable at older ages than at younger ages because of the much smaller absolute number of deaths at older ones. 

In 	\cite{Brouhns2002a} a maximum likelihood estimation based on a Poisson  death count $D^{(i)}_{x,t}$ is proposed to allow heteroskedasticity. In this case, the ILC model for multiple populations reads 
\begin{equation}
\label{eq:poisson_assumption}
D^{(i)}_{x,t}  \sim Poisson (E^{(i)}_{x,t} m^{(i)}_{x,t}) \quad with \quad m^{(i)}_{x,t} = e^{a^{(i)}_x + b^{(i)}_x k^{(i)}_t}
\end{equation}
where $E^{(i)}_{x,t}$ is the number of exposure-to-risk in age $x$ at time $t$ in the population $i$ and the constraints in  \eqref{eq:constraints} still hold $\forall i \in \mathcal{I}$. 
The model parameters can be estimated by solving
\begin{equation}
\argminA_{(a^{(i)}_x)_x, (b^{(i)}_x)_x, (k^{(i)}_t)_t} \sum_{x \in \mathcal{X}} \sum_{t \in \mathcal{T}} \bigg(D^{(i)}_{x,t} (a^{(i)}_x + b^{(i)}_x k^{(i)}_t) - E^{(i)}_{x,t} e^{a^{(i)}_x +b^{(i)}_x k^{(i)}_t} \bigg) + c_i, \quad \forall i \in \mathcal{I}\\
\end{equation}
 where $c_i \in \mathbb{R}$ is a constant which only depends on the data. 
The meaning of the parameters is essentially the same of the corresponding
parameters in the classical LC model. Furthermore, in \cite{Brouhns2002a} the authors do not modify the time-series part of the LC method.






\section{Feed-Forward Neural Networks }
\label{sec:NN}
Feed-forward neural networks are popular methods in data science and machine learning. They can be considered as high-dimensional non-linear regression models and are achieving excellent performance in several fields.  
Feed-forward neural network consists of a set of (non-linear) functions, called units, arranged in layers (input, output and hidden layers) which process and transform data to perform a specific task. 
How the units are connected configures different types of neural networks. A brief description of the neural network blocks used in the paper is provided below.


\subsection{Fully-Connected Neural Networks}
\label{sec:FCN}
Fully-Connected Networks (FCN)  are probably the most popular type of feed-forward neural networks. In FCN,  each unit of a layer is connected to every part of the previous one. 
First,  we describe a single  FCN layer.

Let $\boldsymbol{x} = (x_1, x_2, \dots ,  x_d)^\top\in \R^{q_0}$ be a $q_0$-dimensional input vector, a FCN layer with $q_1\in\N$ hidden units is a function that maps the input $\boldsymbol{x}$ to a new $q_1$-dimensional space 
\begin{equation}
\label{FCN layer mapping}
	\bz : \mathbb{R}^{q_0} \to \mathbb{R}^{q_1}, \quad \quad \bx \mapsto  \bz(\bx) = \left( z_1(\bx), z_2 (\bx), \dots , z_{q_1}(\bx) \right)^\top. 
\end{equation}
Each new feature component $z_j(\boldsymbol{x})$ is a non-linear function of $\bx$
\begin{equation}\label{FCN layer 2}
\bx~\mapsto~
\bz_j(\boldsymbol{x})  =\phi \left( w_{j,0} + \sum_{l = 1}^{q_0} w_{j,l} x_l \right)  = \phi \left( w_{j,0} +  \left\langle \bw_j,  \bx \right\rangle \right), \quad \quad j = 1,  \dots , q_1 , 
\end{equation}
where   $\phi : \mathbb{R} \to \mathbb{R}$ is a (non-linear) activation function,$w_{j,l} \in \mathbb{R}$ represent the network  parameters and $\langle \cdot, \cdot \rangle$ denotes the scalar product in $\mathbb{R}^{q_0}$.

When the FCN is shallow, it presents a single hidden layer followed by the output layer. Differently, a deep FCN provides several stacked  FCN layers and the output of each layer becomes the input of the next one and so for the following layers.
Let $\boldsymbol{q} = \{q_k\}_{1 \leq k \leq m} \in \mathbb{N}^{m}$ be a sequence of integers  defining the size of each layer where $m \in \mathbb{N}$ is the number of hidden layers also called \emph{depth}. 
A deep FCN can be formalised as:
\begin{equation}\label{deep FCN}
\bx ~ \mapsto ~ \bz^{(m:1)} (\bx)=\left( \bz^{(m)} \circ \cdots \circ \bz^{(1)} \right) (\bx)~\in ~\R^{q_m},
\end{equation}
where all mappings $\bz^{(k)}:\mathbb{R}^{q_{k-1}} \to \mathbb{R}^{q_{k}}$ adopt the structure in  \eqref{FCN layer mapping} with weights 
$W^{(k)} = (\bw^{(k)}_j)_{1\leq j \leq q_k} \in \R^{q_k \times q_{k-1}}$ and biases $\bw^{(k)}_0 \in \R^{q_k}$, for $1 \leq k \leq m$.  $\phi_k$ are the activation functions of each layer which could also differ from each other. 

Both shallow and deep FCNs include a final output layer that computes the variable of interest $y$ as a function of the features extracted of the last hidden layer $\bz^{(m:1)} (\bx)$. This layer is a mapping $g:\mathbb{R}^{q_m} \to \mathcal{Y} $ that must be  chosen according to the domain of the  response variable.

Given a specific prediction task, the performance of a deep FCN strongly depends on the  weights $w^{(k)}_{j.l}$ that must be properly calibrated.
Given a specific loss function $L(g(\bz^{(m:1)} (\bx)),y)$, which measures the quality of the predictions produced by the network $g(\bz^{(m:1)} (\bx))$ against the observed values $y$ of the response variable $Y$, network training (or fitting) consists in finding the weights that minimise $L(g(\bz^{(m:1)} (\bx)),y)$.
 It is generally performed via the Back-Propagation (BP) algorithm where the weights are updated iteratively to step-wise decrease the objective function, with each update of the weights based on the gradient of the loss function. An extensive description of network fitting and the BP algorithm is found in 
\cite{Goodfellow2016}.

\subsection{Locally-Connected Neural Networks}
\label{sec:LCN}
The traditional FCN layers do not take into account the spatial structure of the data since they treat elements of the input data that are far or close to each other without distinction. 
The locally-connected network (LCN) layers overcame this problem. They are characterised by the \emph{local connectivity} since each unit of the layer is connected only with a local area of the input, called \emph{receptive field}.
The  resulting weight matrices (called \emph{filters}) present a smaller  size than the input data and  the features extracted  are  functions of a small part of the input data. 
This  induces a significant reduction of the parameters to optimise with respect to the  traditional FCN layers. 
The feature map extracted by a LCN layer depends, among the other hyper-parameters, on the kernel filter' size $m \in \N$ and the stride $s \in \N$. 
The kernel size refers to the dimension of the network filters and determines  the number of parameters while the stride defines the distance between two adjacent receptive fields.
In the standard setting,  the LCN layers use $s = 1$ however, this choice induces  a big overlap between adjacent  receptive fields and much information is repeated. 
Alternatively, a stride $s>1$ could be considered. It would reduce the overlap of receptive fields and lead to computational benefits. 
 Typically, LCNs work on tensors and the locally-connectivity can also be applied to multiple dimensions. 
 For simplicity, we only describe a 1-dimensional LCN layer where $m,s \in \mathbb{N}$  are suitable chosen such that $(d-m)/s  \in \mathbb{N}$.
  
 Let $\bx_i \in \R^{b}, \ 1 \leq i \leq d$ be an ordered sequence of data input and $m,s \in \mathbb{N}$.
 A 1d locally-connected layer with $q \in \N$ filters is a mapping 
 \begin{equation}\label{eq:formal lcn layer}
	\bz : \mathbb{R}^{1 \times d \times b} \to \mathbb{R}^{((d-m)/s +1) \times q },
\bx=(\bx_1,\ldots, \bx_{d}) \mapsto \bz(\bx) = \left(z^{(j)}_{k}(\bx)\right)_{1\le k \le ((d-m)/s+1), 1\le j\le q}.
\end{equation}
We keep the ``1'' in the following notation to highlight that this is a 1d-LCN layer.
Denoting by $W^{(j)}_k = (\bw^{(j)}_{k, 1}, \bw^{(j)}_{k, 2}, \dots, \bw^{(j)}_{k, m}) \in \mathbb{R}^{b \times m}$ the kernel 
 filters  and $w^{(j)}_{k, 0} \in \mathbb{R}$ the bias terms for $ k = 1, \dots, ((d-m)/s +1)$ and $ j = 1, \dots, q$, each component of the new mapping can be espressed as 
\begin{equation}\label{explanation of lcn}
\bx ~\mapsto~
z^{(j)}_{k} =z^{(j)}_{k}(\bx) =   \phi \left( w^{(j)}_{k, 0}+ \sum_{l=1}^m \langle \bw^{(j)}_{k, l}, \bx_{m+1+(k-1)\cdot s-l} \rangle\right). 
\end{equation}
Unlike those  obtained using a FCN layer, the features extracted from a LCN layer are  functions of only a small part of the input data.

\subsection{Convolutional Neural Network}
\label{sec:CNN}
Convolutional neural networks (CNNs) introduced by \cite{LeCun1990} are  variants of the locally-connected neural networks. In fact, they result even more popular than the previous ones owing to the impressive results achieved in  image recognition and time-series forecasting tasks.
In addition to sparse connectivity, CNN layers exploit  the \emph{parameter sharing}. Indeed, while in the LCN layer each filter is used exactly once,  the CNN layer is based on the idea  that the same filter can be used to compute many different features.
More in detail, the same filter slides along the input surface and, multiplying different receptive fields, extracts different new features. 
This means that, instead of learning a separate set of parameters for every location,  only one set of parameters must be calibrated,  inducing a further reduction of the parameters to learn with respect to the LCN layers. 
Furthermore, similarly to LCNs, kernel size and stride are two hyper-parameters which influence the features map extracted from a CNN layer.

In notation, a CNN layer uses $W^{(j)} = W^{(j)}_{k}, \ \forall k: 1 \leq k \leq ((d-m)/s +1)$ for each  $j = 1, 2, \dots, q$. It is a mapping with the same structure of \eqref{eq:formal lcn layer} and  each component is given by
\begin{equation}\label{explanation of convolutional}
\bx ~\mapsto~
z^{(j)}_{k} =z^{(j)}_{k}(\bx) =  \phi \left( w^{(j)}_{ 0}+ \sum_{l=1}^m \langle \bw^{(j)}_{l}, \bx_{m+1+(k-1)\cdot s-l} \rangle\right).
\end{equation}
Also in this case, the features extracted by a CNN layer are  functions of only a small part of the input data. A detailed description of the CNN layers and a comparative discussion against the LCN layers can be found in \cite{Goodfellow2016}. 

\subsection{Embedding network}
\label{Embedding layers}
Data often present categorical variables. Examples in mortality modeling context are the region $r \in \mathcal{R}$ and the gender $g \in \mathcal{R}$ to which a particular mortality rate refers. Dummy coding or one-hot encoding are popular approaches to deal with categorical variables among statisticians and the machine learning community, respectively. However, when observations present many categorical variables or when the variables have many different labels, these coding schemes produce high dimensional sparse vectors, which often causes computational and calibration difficulties.

Embedding layers provide an elegant way to deal with categorical input variables.
They  are already intensively  employed in  Natural Language Processing  \cite{Bengio2003} and recently are introduced in the actuarial literature by Richman, see \cite{Richman2020a, Richman2020b}.
In essence, they allow to learn a low-dimensional representation of a categorical variable. Thereby,  every level of the considered categorical variable is mapped to a vector in $\mathbb{R}^{q_\mathcal{P}}$ for some $q_\mathcal{P} \in \mathbb{N}$. These vectors are then simply parameters of the neural network that have to be trained \cite{Guo2016}.
In the new space learned by the embedding layer, labels similar for the  task of interest present a small euclidean distance while different labels present a larger one.

Formally, let $\mathcal{P} = \{p_1, p_2,\dots, p_{n_\mathcal{P}} \}$ be the (finite) set of categories of the qualitative variable and $n_\mathcal{P} = |\mathcal{P}|$ be its the cardinality.  An embedding layer is a mapping
\begin{align*}
\bz_\mathcal{P} : \mathcal{P} \to \mathbb{R}^{q_\mathcal{P} }
, \quad \quad p \mapsto  \bz_\mathcal{P}(p),
\end{align*}

where $q_\mathcal{P} \in \mathbb{N}$ is a hyper-parameter denoting the size of the embedding layer. 
The number of embedding weights that must be learned during training is $n_\mathcal{P} q_\mathcal{P}$ and the embedding size is typically   $q_\mathcal{P} \ll n_\mathcal{P}$.

\section{The neural network approach for ILC  models fitting}
\label{sec:core}


In this section, we introduce the proposed general neural network architecture for the ILC models fitting. In this regard, we keep the same notation introduced in Section \ref{sec:LCmodel} and consider different populations indexed by $i \in \mathcal{I}$,  where populations may differ in gender $g \in \mathcal{G}= \{{\tt male},  {\tt female}\}$ and country $r \in \mathcal{R}$ such that $i = (r,g) \in \mathcal{I} = \mathcal{R} \times \mathcal{G}$. 

\subsection{Model formalisation}
We develop  a neural network  which models the mortality of a large set of populations by replicating  the ILC models' structure. 
The mortality of each population is modeled by an own LC model; however, unlike the standard approach, the model fitting is performed in a single stage using all available mortality data. 
Each mortality experience processed by the network consists of the curve of the log-mortality rates for all ages $log (\boldsymbol{m}_t^{(i)}) = \big( log(m_{x,t}^{(i)} \big)_{x \in \mathcal{X}}$  and the gender and region labels, respectively $r \in \mathcal{R}$ and $g\in \mathcal{G}$, that identify uniquely the population $i = (r, g) \in \mathcal{I}$. 

The network architecture, based on the blocks described in Section \ref{sec:NN}, can be conceptually divided into three subnets which process the data inputs separately. Each  one of these subnets aims to extract one component  of the model in \eqref{eq:ILC}. 
More in details, the outputs of first two subnets substitute  $\boldsymbol{a}^{(i)} = \big( a_x^{(i)} \big)_{x \in \mathcal{X}} \in \mathbb{R}^{|\mathcal{X}|}$ and $\boldsymbol{b}^{(i)} = \big( b_x^{(i)} \big)_{x \in \mathcal{X}} \in \mathbb{R}^{|\mathcal{X}|}$. Since these LC parameters depend only on the population $i$, these subnets process only the region and gender labels. The outputs of these two subnets are two time-independent vectors that present as many components as the ages considered.
The third subnet aims to extract the factor $k_t^{(i)} \in \mathbb{R}$, which summaries the mortality dynamics at time $t$ in the population $i$. In this case, the subnet takes as input the curve of log-mortality rates $log(\mathbf{m}_t^{(i)})\in \mathbb{R}^{|\mathcal{X}|}$ in the population $i$ at time $t$ and produces as output a single  value. In other words, this subnet encodes the curve of the log-mortality rates into a real value. 
Finally, the extracted factors  are  combined to provide an approximation of the log-mortality rates' curve using the functional form of the LC model.

\begin{figure}[htb!]
	\centering
	\includegraphics[width=1\linewidth, scale = 0.5]{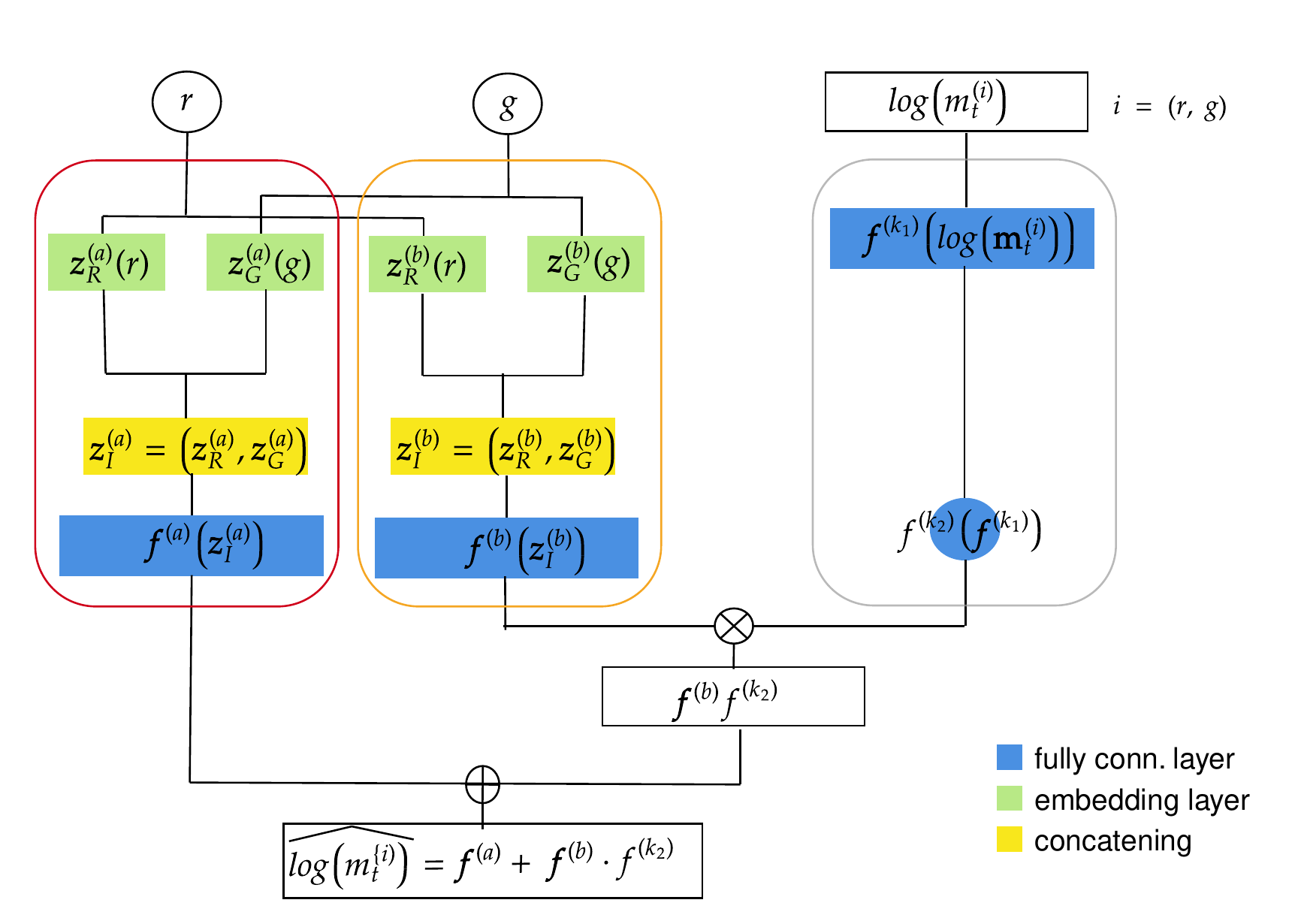}
    \caption{Graphical representation of the neural network architecture for ILC models fitting.}
    \label{fig:model_diagram}
\end{figure}

A pictorial representation of the network architecture is illustrated in Figure \ref{fig:model_diagram}, while a formal description is provided below.

In details, the first subnet (confined within the red diagram) consists of two embedding layers and a FCN layer.
Formally, let $q^{(a)}_{\mathcal{R}}, q^{(a)}_{\mathcal{G}} \in \mathbb{N}$ be the hyper-parameter values defining the size of the two embedding layers,  they  map $r \in \mathcal{R}$ and $g \in \mathcal{G}$ into real-valued vectors:
\begin{eqnarray*}
	\bz_{\cal R}^{(a)} : \mathcal{R} \to \mathbb{R}^{q^{(a)}_{\cal R}}, &&\quad \quad r \mapsto  \bz^{(a)}_{\cal R}(r) =\left( z^{(a)}_{{\cal R},1}(r), z^{(a)}_{{\cal R},2} (r), \dots , z^{(a)}_{{\cal R},q^{(a)}_{\cal R}}(r) \right)^\top, 
\\	
	\bz_{\cal G}^{(a)} : \mathcal{G} \to \mathbb{R}^{q^{(a)}_{\cal G}}, &&\quad \quad g \mapsto  \bz^{(a)}_{\cal G}(g) =\left( z^{(a)}_{{\cal G},1}(g), z^{(a)}_{{\cal G},2} (g), \dots , z^{(a)}_{{\cal G},q^{(a)}_{\cal G}}(g) \right)^\top.
\end{eqnarray*}
Since $\bz_{\mathcal{R}}^{(a)}(r)$  is a new set of features representing  the region $r$ and  $\bz_{\mathcal{G}}^{(a)}(g)$  is a new representation of the gender $g$, 
the vector $\bz_{\mathcal{I}}^{(a)} = \bz_{\mathcal{I}}^{(a)} (r, g)  =  \big(\big(\bz_{\mathcal{R}}^{(a)}(r)\big)^\top,\big(\bz_{\mathcal{G}}^{(a)}(g) \big)^\top\big)^\top\in \mathbb{R}^{q^{(a)}_{\mathcal{I}}} $ (with $q_{\mathcal{I}}^{(a)} = q_{\mathcal{R}}^{(a)} +q_{\mathcal{G}}^{(a)}$), obtained concatenating the output of these two embedding layers,  can be  understood as a learned representation of the population $i = (r, g)$.  It is then processed by a FCN layer which provides as many units as the  age considered. 
This layer maps $\bz_{\mathcal{I}}^{(a)}$ in a new $|\mathcal{X}|$-dimensional real-valued space
\begin{equation*}
	\bff^{(a)} : \mathbb{R}^{q^{(a)}_{\mathcal{I}}} \to \mathbb{R}^{|\mathcal{X}|}, \quad \quad \bz^{(a)}_{\mathcal{I}} \mapsto  \bff^{(a)}(\bz^{(a)}_{\mathcal{I}}) = \left( f^{(a)}_{x_{0}}(\bz^{(a)}_{\mathcal{I}}), f^{(a)}_{x_{1}} (\bz^{(a)}_{\mathcal{I}}), \dots , f^{(a)}_{x_{\omega}}(\bz^{(a)}_{\mathcal{I}}) \right)^\top.
\end{equation*}
Each new feature  $f^{(a)}_x(\bz^{(a)}_{\mathcal{I}})$ is a age-specific function of the vector $\bz^{(a)}_{\mathcal{I}}$
\begin{equation}\label{FCN layer ax}
\bz^{(a)}_{\mathcal{I}}~\mapsto~
f^{(a)}_x(\bz^{(a)}_{\mathcal{I}})  =\phi^{(a)} \bigg( w^{(a)}_{x,0} + \sum_{l = 1}^{q_{\mathcal{I}}^{(a)}} w^{(a)}_{x,l} z^{(a)}_{\mathcal{I}, l} \bigg)= \phi^{(a)} \left( w^{(a)}_{x,0} +  \left\langle \bw^{(a)}_x,  \bz^{(a)}_{\mathcal{I}} \right\rangle \right), \quad  x \in \mathcal{X},
\end{equation}
where   $\phi^{(a)} : \mathbb{R} \to \mathbb{R}$ is a (non-linear) activation function, $w^{(a)}_{x,l} \in \mathbb{R}$ are the network  parameters.

The second subnet (confined within the orange diagram) presents the same architecture of the first one. 
It provides two embedding layers of size $q^{(b)}_{\mathcal{R}}, q^{(b)}_{\mathcal{G}} \in \mathbb{N}$  described as
\begin{eqnarray*}
	\bz_{\cal R}^{(b)} : \mathcal{R} \to \mathbb{R}^{q^{(b)}_{\cal R}}, &&\quad \quad r \mapsto  \bz^{(b)}_{\cal R}(r) =\left( z^{(b)}_{{\cal R},1}(r), z^{(b)}_{{\cal R},2} (r), \dots , z^{(b)}_{{\cal R},q^{(b)}_{\cal R}}(r) \right)^\top, 
\\
		\bz_{\cal G}^{(b)} : \mathcal{G} \to \mathbb{R}^{q^{(b)}_{\cal G}}, &&\quad \quad g \mapsto  \bz^{(b)}_{\cal G}(g)  =\left( z^{(b)}_{{\cal G},1}(g), z^{(b)}_{{\cal G},2} (g), \dots , z^{(b)}_{{\cal G},q^{(b)}_{\cal G}}(g) \right)^\top,
\end{eqnarray*}
and a $|\mathcal{X}|$-dimensional FCN layer which takes  the vector  $\bz_{\mathcal{I}}^{(b)} = \bz_{\mathcal{I}}^{(b)} (r, g)  =  \big(\big(\bz_{\mathcal{R}}^{(b)}(r)\big)^\top,\big(\bz_{\mathcal{G}}^{(b)}(g)\big)^\top \big)^\top\in \mathbb{R}^{q^{(b)}_{\mathcal{I}}} $ (with $q_{\mathcal{I}}^{(b)} = q_{\mathcal{R}}^{(b)} +q_{\mathcal{G}}^{(b)}$) as input.   This layer maps $\bz_{\mathcal{I}}^{(b)}$ in $|\mathcal{X}|$-dimensional real-valued space
\begin{equation*}\label{FCN layer}
	\bff^{(b)} : \mathbb{R}^{q^{(b)}_{\mathcal{I}}} \to \mathbb{R}^{|\mathcal{X}|}, \quad \quad \bz^{(b)}_{\mathcal{I}} \mapsto  \bff^{(b)}(\bz^{(b)}_{\mathcal{I}}) = \left( f^{(b)}_{x_{0}}(\bz^{(b)}_{\mathcal{I}}), f^{(b)}_{x_{1}} (\bz^{(b)}_{\mathcal{I}}), \dots , f^{(b)}_{x_{\omega}}(\bz^{(b)}_{\mathcal{I}}) \right)^\top .
\end{equation*}
Also in this case, each new  component $f^{(b)}_x(\bz^{(b)}_{\mathcal{I}})$ is an age-specific  function of $\bz^{(b)}_{\mathcal{I}}$
\begin{equation}\label{FCN layer bx}
\bz^{(b)}_{\mathcal{I}}~\mapsto~
f^{(b)}_x(\bz^{(j)}_{\mathcal{I}})  =\phi^{(b)} \bigg( w^{(b)}_{x,0} + \sum_{l = 1}^{q_{\mathcal{I}}^{(b)}} w^{(b)}_{x,l} z^{(b)}_{\mathcal{I}, l} \bigg)  = \phi^{(b)} \bigg( w^{(b)}_{x,0} +  \left\langle \bw^{(b)}_x,  \bz^{(b)}_{\mathcal{I}} \right\rangle \bigg), \quad  x \in \mathcal{X}, 
\end{equation}
with  $\phi^{(b)} : \mathbb{R} \to \mathbb{R}$ and $w^{(b)}_{x,l} \in \mathbb{R}$.

It is important to remark that, despite the first two subnets present the same architecture,  the weights to learn in each layer are different.
Denoting by $\bw^{(j)}_{0} = (w^{(j)}_{x,0} )_{x \in \mathcal{X}} \in \mathbb{R}^{\mid \mathcal{X} \mid }$ and
$W^{(j)} = (\bw^{(j)}_{x,\mathcal{I}} )^\top_{x \in \mathcal{X}}  \in \mathbb{R}^{ \mid \mathcal{X} \mid \times q^{(j)}_{\mathcal{I}}}$, $\forall j \in \{a,b\}$, the output of the first two subnets can be written in compact form  
\begin{equation}
\label{eq:ax_vectorial}
 \bff^{(a)} \big(  \bz_{\mathcal{I}}^{(a)} \big)   =  \phi^{(a)} \bigg( \bw^{(a)}_{0} +  \left\langle W^{(a)}, \bz_{\mathcal{I}}^{(a)} \right\rangle  \bigg)
 = \phi^{(a)} \bigg( \bw_{0}^{(a)}+\left\langle W_{\mathcal{R}}^{(a)} , \bz_{\mathcal{R}}^{(a)} (r) \right\rangle + \left\langle W_{\mathcal{G}}^{(a)} , \bz_{\mathcal{G}}^{(a)} (g) \right\rangle \bigg), 
\end{equation}
\begin{equation}
\label{eq:bx_vectorial}
 \bff^{(b)} \big(  \bz_{\mathcal{I}}^{(b)} \big)   =  \phi^{(b)} \bigg( \bw^{(b)}_{0} +  \left\langle W^{(b)}, \bz_{\mathcal{I}}^{(b)} \right\rangle  \bigg)
 = \phi^{(b)} \bigg( \bw_{0}^{(b)}+\left\langle W_{\mathcal{R}}^{(b)} , \bz_{\mathcal{R}}^{(b)} (r) \right\rangle + \left\langle W_{\mathcal{G}}^{(b)} , \bz_{\mathcal{G}}^{(b)} (g) \right\rangle \bigg), 
\end{equation}
where one could carry out the decomposition $W^{(j)}   = \big( W_{\mathcal{R}}^{(j)}, W_{\mathcal{G}}^{(j)} \big)$  of the   matrices of the FCN layers to distinguish the weights which refer to  the gender-specific and the region-specific features.  
 
The architecture of the third subnet (inside the gray diagram) is different from the previous ones. It consists of some staked feed-forward neural network layers which process the curve of log-mortality rates. In this case, a large discretionary in the number and type of neural network layers to employ is left to the modeler. 
For sake of simplicity, we described the model  considering two FCN layers and  postponed a  comparative discussion with the LCN and CNN layers to the numerical experiments' section.
 Let $q_{z_1}\in \mathbb{N}$ and $ q_{z_2} = 1$ be two hyper-parameters defining the size of two FCN layers.  The first FCN layer maps $log(\boldsymbol{m}_{t}^{(i)})$ into a $q_{z_1}$-dimensional real-valued space:
 \begin{equation*}
	\bff^{(k_1)} : \mathbb{R}^{|\mathcal{X}|} \to \mathbb{R}^{q_{z_1}}, \quad \quad log(\boldsymbol{m}_{t}^{(i)}) \mapsto  \bff^{(k_1)} \big(log(\boldsymbol{m}_{t}^{(i)}) \big) = \left( f^{(k_1)}_1\big(log(\boldsymbol{m}_{t}^{(i)})\big), \dots , f^{(k_1)}_{q_{z_1}}\big(log(\boldsymbol{m}_{t}^{(i)}) \big) \right)^\top,
\end{equation*}
where each new feature component $f^{(k_1)}_s(log(\boldsymbol{m}_{t}^{(i)}))$ is function of the mortality rates of all ages 
\begin{equation}\label{FCN layer kt1}
log(\boldsymbol{m}_{t}^{(i)})~\mapsto~
f^{(k_1)}_s\big(log(\boldsymbol{m}_{t}^{(i)})\big)  =\phi^{(k_1)} \left( w^{(k_1)}_{s,0} +  \left\langle \bw^{(k_1)}_s,  log(\boldsymbol{m}_{t}^{(i)}) \right\rangle \right), \quad  s = 1,  \dots , q_{z_1},
\end{equation}
where $w^{(k_1)}_{s,0} \in \R$ and $\bw^{(k_1)}_s \in \mathbb{R}^{|\mathcal{X}|}$ are parameters.
Otherwise, by using LCN layer we would obtain new features which are functions of a segment of the log-mortality curve.
The second FCN layer of size $q_{z_2} = 1$ is a mapping
  \begin{equation*}\label{eq:2FCN_3net}
	f^{(k_2)} : \mathbb{R}^{q_{z_1}} \to \mathbb{R}, \quad \quad \bff^{(k_1)} \big( log(\boldsymbol{m}_{t}^{(i)}) \big) \mapsto  f^{(k_2)} \big(\bff^{(k_1)} \big(log(\boldsymbol{m}_{t}^{(i)}) \big)\big) =   \big(f^{(k_2)} \circ \bff^{(k_1)}\big) \big(log(\boldsymbol{m}_{t}^{(i)})\big) . 
\end{equation*}
It extracts a single new feature 
 \begin{equation}
\label{FCN layer kt2}
   (f^{(k_2)} \circ \bff^{(k_1)}) (log(\boldsymbol{m}_{t}^{(i)})) = \phi^{(k_2)} \bigg( w^{(k_2)}_0 +  \left\langle \bw^{(k_2)}, \phi^{(k_1)} \bigg(  \bw^{(k_1)}_0 +  \left\langle W^{(k_1)}, log(\boldsymbol{m}_{t}^{(i)})\right\rangle \bigg) \right\rangle \bigg),
\end{equation}
 where $w_0^{(k_2)} \in \mathbb{R}, \bw_0^{(k_1)} = (w_{s,0}^{(k_1)})_{1\leq s \leq q_{z_1}}\in \mathbb{R}^{q_{z_1}}, \bw^{(k_2)}\in \mathbb{R}^{q_{z_1}}, W^{(k_1)}  = (\bw_s^{(k_1)} )^{\top}_{1\leq s \leq {q_{z_1}}}\in \mathbb{R}^{q_{z_1} \times \mid \mathcal{X} \mid}$ are network parameters and  $\phi^{(j)}(\cdot) : \mathbb{R} \to \mathbb{R}$ for $j \in \{k_1, k_2\}$ are activation functions.
Basically, the first FCN layer encodes the log-mortality curve into a $q_{z_1}$- dimensional real-valued vector, the second layer further compresses this results into in a single real value.
 
Finally, an approximation  of log-mortality curve   at time $t$ in the population $i$ can be obtained as
\begin{equation}
\begin{split}
\label{eq:full_model_compressed_matrix}
  \widehat{ log(\boldsymbol{m}_{t}^{(i)})}=  \bff^{(a)} \big(  \bz_{\mathcal{I}}^{(a)} \big)  +   \bff^{(b)} \big(  \bz_{\mathcal{I}}^{(b)} \big)   
   (f^{(k_2)} \circ \bff^{(k_1)}) (log(\boldsymbol{m}_{t}^{(i)}))
\end{split}
\end{equation}
where each age component is given by
\begin{equation}
\begin{split}
\label{eq:full_model_compressed}
  \widehat{ log({m}_{x,t}^{(i)})}=  f_x^{(a)} \big(  \bz_{\mathcal{I}}^{(a)} \big)  +   f_x^{(b)} \big(  \bz_{\mathcal{I}}^{(b)} \big)   
   \big(f^{(k_2)} \circ \bff^{(k_1)}\big) \big(log(\boldsymbol{m}_{t}^{(i)})\big).
\end{split}
\end{equation}
A simple interpretation of all terms in (\ref{eq:full_model_compressed})  can be provided:
\begin{itemize}
    \item $ f_x^{(a)} \big(  \bz_{\mathcal{I}}^{(a)} \big)\in \mathbb{R}$ is a population and age-specific term that plays the same role of  ${a}_x^{(i)}$ in the LC model.
    \item $ f_x^{(b)} \big(  \bz_{\mathcal{I}}^{(b)} \big) \in \mathbb{R}$ is a population and age-specific term that plays the same role of  $b_x^{(i)}$ in the LC model. 
    \item $ (f^{(k_2)} \circ \bff^{(k_1)}) (log(\boldsymbol{m}_{t}^{(i)})) \in \mathbb{R}$ is a population and time-specific term that plays the same role of the $k_t^{(i)}$ in the LC model. 
\end{itemize}

 In addition,  setting linear activation $\phi^{(j)}(x)=x, \ \forall j \in \{a,b\}$ and  expanding all the terms in \eqref{eq:full_model_compressed}, the model can be written as
\begin{equation*}
\label{eq:full_model}
  \widehat{ log({m}_{x, t}^{(i)})}=  \bigg( w^{(a)}_{x,0} +  \left\langle \bw_x^{(a)}, \bz_{\mathcal{I}}^{(a)} \right\rangle  \bigg)\  +   \end{equation*}
  \begin{equation}
+ \bigg( w^{(b)}_{x,0} +  \left\langle \bw_x^{(b)}, \bz_{\mathcal{I}}^{(b)} \right\rangle \bigg) \cdot \phi^{(k_2)} \bigg( w^{(k_2)}_0 +  \left\langle \bw^{(k_2)}, \phi^{(k_1)} \bigg(  \bw^{(k_1)}_0 +  \left\langle W^{(k_1)}, log(\boldsymbol{m}_{x}^{(i)}) \right\rangle  \bigg) \right\rangle \bigg).
\end{equation}
In this case, some further interpretations can be argued. 
The  term $\bigg( w^{(a)}_{x,0} +  \left\langle \bw_x^{(a)}, \bz_{\mathcal{I}}^{(a)} \right\rangle  \bigg)$ can be further decomposed into ${w}^{(a)}_{x, 0}$, which can be interpreted as a  population-independent $a_x$ parameter, and $ \left\langle \bw_x^{(a)}, \bz_{\mathcal{I}}^{(a)}\right\rangle =\left\langle \bw_{x,\mathcal{R}}^{(a)}, \bz_{\mathcal{R}}^{(a)}(r)\right\rangle+\left\langle\bw_{x,\mathcal{G}}^{(a)}, \bz_{\mathcal{G}}^{(a)}(g) \right\rangle$, which can be interpreted as a population-specific $a_x$ correction. In particular, it is the sum  of the region-specific correction term $\left\langle\bw_{x,\mathcal{R}}^{(a)}, \bz_{\mathcal{R}}^{(a)} (r)\right\rangle$ and the gender-specific  correction term $\left\langle\bw_{x,\mathcal{G}}^{(a)},  \bz_{\mathcal{G}}^{(a)} (r)\right\rangle$.
The same decomposition can be applied  to $\bigg( w^{(b)}_{x,0} +  \left\langle \bw_x^{(b)}, \bz_{\mathcal{I}}^{(b)} \right\rangle \bigg)$.

 \subsection{Model Fitting and Forecasting}
 
 As anticipated in the previous sections, the neural network model's performance depends on the network parameters that must be appropriately calibrated.
Denoting by $\psi$ the full set of the   network model's parameters  described above, it  can be splitted into two groups.
The first group concerns the population-specific parameters, namely the embedding parameters, $\bz^{(a)}_{\cal R}(r), \bz^{(b)}_{\cal R}(r),\ \forall r \in \mathcal{R}$ and  $\bz^{(a)}_{\cal G}(g), \bz^{(b)}_{\cal G}(g),\ \forall g \in \mathcal{G}$ which contribute only to the population-specific  LC model. 
The remaining $\bw^{(j)}_{0},  W^{(j)},\ \forall j \in \{a,b, k_1\}$ and $\bw^{(k_2)}, w^{(k_2)}_0$ are cross-population parameters which contribute to all the individual LC models.  
Choosing a specific loss function, these parameters are iteratively  adjusted via BP algorithm to pursue its minimum.
During the training,  we also apply the dropout \cite{srivastava2014dropout} in some layers of the networks  to regularise.
Dropout refers to ignoring  some random chosen units  during the network fitting.
This technique forces a neural network to learn more robust features and prevent over-fitting. 
Once the model training was completed and an estimation of the optimal set of parameters $\hat{\psi}$ was obtained, population-specific parameters and cross-population parameters can be used to compute the  terms in the model presented in \eqref{eq:full_model_compressed}. 
Since these quantities have the  same meaning as the LC model terms,  we could consider them as Neural Network (NN) estimates of the LC parameters:
\begin{equation}
\label{eq:ax}
    \hat{a}_{x, NN}^{(i)} =   \phi^{(a)} \bigg(  \hat{w}_{x, 0}^{(a)}+\left\langle \hat{\bw}_{x,\mathcal{R}}^{(a)},  \hat{\bz}_{\mathcal{R}}^{(a)} (r) \right\rangle + \left\langle \hat{\bw}_{x,\mathcal{G}}^{(a)},  \hat{\bz}_{\mathcal{G}}^{(a)} (g) \right\rangle \bigg),  \quad \quad \forall x \in \mathcal{X}, \forall i \in \mathcal{I},
\end{equation}
\begin{equation}
\label{eq:bx}
    \hat{b}_{x, NN}^{(i)} =  \phi^{(b)} \bigg(  \hat{w}_{x, 0}^{(b)}+\left\langle \hat{\bw}_{x,\mathcal{R}}^{(b)},  \hat{\bz}_{\mathcal{R}}^{(b)} (r) \right\rangle + \left\langle \hat{\bw}_{x,\mathcal{G}}^{(b)},  \hat{\bz}_{\mathcal{G}}^{(b)} (g) \right\rangle \bigg), \quad \quad \forall x \in \mathcal{X}, \forall i \in \mathcal{I},
\end{equation}
\begin{equation}
\label{eq:kt}
    \hat{k}_{t, NN}^{(i)} =   \phi^{(k_2)} \bigg( \hat{w}^{(k_2)}_0 +  \left\langle \hat{\bw}^{(k_2)}, \phi^{(k_1)} \bigg(  \hat{\bw}^{(k_1)}_0 +  \left\langle \hat{W}^{(k_1)}, log(\boldsymbol{m}_{x}^{(i)}) \right\rangle  \bigg) \right\rangle \bigg),\quad \quad \forall t \in \mathcal{T}, \forall i \in \mathcal{I}. 
\end{equation}

Also in this case, to avoid identifiability problems, the constraints in \eqref{eq:constraints} are imposed. 
We do not modify the time-series part of the ILC approach; forecasting is performed assuming that $\hat{a}_{x, NN}^{(i)}$ and $\hat{b}_{x, NN}^{(i)}$  are constant over time while $\hat{k}_{t, NN}^{(i)}$ is projected  with a random walk with drift, $ \forall i \in \mathcal{I}$.
It is interesting to note that often the number of LC parameters $(a_x^{(i)})_x, (b_x^{(i)})_x, (k_t^{(i)})_t, \forall i \in \mathcal{I}$  can be larger than the number of network weights.

\section{Numerical experiments}
\label{sec:numerical_experiments}
In this section, some numerical experiments to validate the proposed approach are conducted. 

The data source is  the Human Mortality Database (HMD), which provides mortality data for male and female populations of a large set of countries. 
Following the experiments' scheme in \cite{Perla2021, Richman2020}, we only consider mortality data from $1950$ onwards, and we set $1999$ as final observation year.

Let $\mathcal{T} = \{  t \in \N:  1950 \leq t <2019\}$ be the full set of available years and  $\mathcal{T}_{1} = \{  t \in \mathcal{T}:  t <2000\}, \mathcal{T}_{2} = \{ t \in \mathcal{T}:  t >2000\}$ such that $\mathcal{T}_{1} \cup \mathcal{T}_{2} = \mathcal{T}$. 
The aim is to forecast, as accurately as possible, the mortality rates of calendar years in $\mathcal{T}_{2}$  using a  model fitted on mortality data of calendar years in $\mathcal{T}_{1}$. Using the machine learning terminology, the mortality rates of calendar years in $\mathcal{T}_{1}$ represent the \emph{training set}, while the mortality rates of calendar years in $\mathcal{T}_{2}$ represent the \emph{testing set}. 

First, a careful round of data pre-processing was carried out. We consider
only  male and female populations of countries for which at least 10 calendar years of mortality data before $1999$ are available. We define $\mathcal{R} $ as the set of selected countries and, in our experiments, we observe that $|\mathcal{R}|= 40$ and $|\mathcal{I}| = 80$ since $\mathcal{I} = \mathcal{R} \times \mathcal{G} = \mathcal{R} \times \{ {\tt male}, {\tt female}\}$. The full list of countries in $\mathcal{R}$ can be found in Table \ref{tab:countries_in_R} in Appendix \ref{app:data_description}. 
Furthermore, only ages in  $\mathcal{X} = \{  x \in \mathbb{Z}_+ : 0 \leq x < 100 \}$ were considered (with $|\mathcal{X}| = 100$) and,  following the scheme adopted in \cite{Perla2021, Richman2020}, mortality rates recorded  as zero or  missing were imputed using the average rate at that age across all countries for that gender in that year. 

After the pre-processing stage, we discuss the neural network models that we experimented.
Since no golden rules exist for the neural network design,  other neural network variants  were compared against the general model described in Section \ref{sec:core}. 
They differ from each other in the  third subnet's design which processes the log-mortality curve $log(\boldsymbol{m}_{x}^{(i)}) $:
\begin{itemize}
    \item The first network, called LC\_FCN, is  the architecture described in Section \ref{sec:core}, which uses two FCN layers.
    We set the size of the two layers respectively $q_{z_1} = 25,  q_{z_2} = 1$.
\item The second variant  replaces the first FCN layer in the third subnet with a LCN layer. In this case, the curve   $log(\boldsymbol{m}_{x}^{(i)}) $   is processed by a LCN layer where the locally-connectivity is applied to the age-related dimension. 
It appears reasonable to believe that a successful features extraction can be obtained by processing separately small segments of the log-mortality curve. 
We set the number of filters $k=1$ and the kernel size and the stride equal to $|\mathcal{X}|\cdot (1/25) =  4$. 
In this setting, there is no overlap between adjacent  receptive fields  and the layer extracts a feature from each  group of 4 adjacent ages ($0-3, 4-7, \dots, 96-99$). The output of this layer is a 25-dimensional vector that is further processed by a FCN as in the LC\_FCN model. 
The adoption of the LCN layer  involves a significant reduction of the parameters to optimise, due to the local connectivity. We call this variant LC\_LCN. 
\item The latest architecture substitutes the first LCN layer of the third subnet with a CNN layer keeping  the same  hyper-parameter setting of the LCN layer-based model. In this case, the features extracted from the different 4-sized age groups are obtained using the same set of parameters. This further reduces the number of parameters to learn with respect to the LCN layer since  the CNNs are characterised by  local connectivity and weight sharing. This variant is named LC\_CNN. 
\end{itemize}

As far as concern the first two subnets, we set the size of the  embedding layers $q^{(a)}_{\mathcal{R}} = q^{(a)}_{\mathcal{G}} = q^{(b)}_{\mathcal{R}} = q^{(b)}_{\mathcal{G}} = 5$ as in  \cite{Perla2021, Richman2020}. 
In addition, we consider only linear  activation $\phi^{(j)}(x) = x \ \forall j \in \{a, b, k_1, k_2 \}$. 
All the analyses are carried out in the {\tt R} environment and the neural network models considered were developed using the {\tt R} package  keras \cite{Chollet2018}. 

Table \ref{tab:parameters} reports the total number of network weights for each neural network model defined above. 
\begin{table}[ht]
\centering
\caption{Number of network parameters for the neural network models considered. }
\medskip
\small
\begin{tabular}{| l|r|}
\hline
Model	&	\# parameters	\\
	\hline
		LC\_FCN	&	5.171	\\
	LC\_LCN	&	2.771  \\
	LC\_CONV	&	2.651	\\
\hline
\end{tabular}
\label{tab:parameters}
\end{table}

We observe that the difference  between the number of parameters of the LC\_FCN and LC\_LCN is quite large. This means that by introducing local connectivity in the first layer of the third subnet, almost  50\%  of the parameters can be saved.
On the contrary, the difference between LC\_LCN and LC\_CNN is quite small. Then, the further reduction of the number of weights induced by the weight sharing mechanism is quite limited.

\subsection{MSE minimisation}
\label{sec:experiments_ols}

In the first stage,  all the network models are fitted minimising the Mean Squared Error (MSE).
The training sample includes 3598 mortality examples which are processed for 2000 epochs.  
The choice of the MSE as loss function can be motivated by arguing that the original paper suggests fitting the LC model using the SVD to perform the PCA.  Since the PCA can be expressed as an optimisation problem, in which the MSE between the original data and the data reconstructed using an approximating linear, the use of the MSE as loss function appears reasonable. 
Furthermore, it is remarkable that the MSE minimisation is equivalent to   the likelihood  maximising in the gaussian assumption of the mortality rates \cite{Richman2020}. 

In this setting, the network training involves the minimisation of the following loss function
\begin{equation*}
L(\psi) =  \sum_{x \in \mathcal{X}} \sum_{i \in \mathcal{I}}  \sum_{t \in \mathcal{T}} \bigg(  log({m}_{x,t}^{(i)}) - \phi^{(a)}\bigg( w^{(a)}_{x,0} +\left\langle \bw_x^{(a)}, \bz_{\mathcal{I}}^{(a)} \right\rangle  \bigg)  +
\end{equation*}
\begin{equation}
-   \phi^{(b)}\bigg( w^{(b)}_{x,0} +  \left\langle \bw_x^{(b)}, \bz_{\mathcal{I}}^{(b)} \right\rangle \bigg) \cdot  \phi^{(k_2)} \bigg( w^{(k_2)}_0 +  \left\langle \bw^{(k_2)}, \phi^{(k_1)} \bigg(  \bw^{(k_1)}_0 +  \left\langle W^{(k_1)}, log(\boldsymbol{m}_{x}^{(i)}) \right\rangle  \bigg) \right\rangle \bigg)\bigg)^2.  \\
\end{equation}
The Adam optimiser algorithm \cite{Kingma2014} with the parameter values taken at the defaults is used. 
When we use the MSE loss function to train the networks, we add the label ``\_mse''  to  of all the neural network  models' names. 

Once the network fitting is completed and the set of the optimal network parameters is estimated, the corresponding NN estimations of the LC parameters   can be computed accordingly with eq. \eqref{eq:ax},  \eqref{eq:bx} and \eqref{eq:kt}\footnote{It should be noted that  \eqref{eq:kt}  expresses  $ \hat{k}_{t, NN}^{(i)}$ for the LC\_FCN model. This formula must be suitably modified according to Sections \ref{sec:LCN} and \ref{sec:CNN} when considering respectively the  LC\_LCN and LC\_CNN models.}. 
Forecasting is performed as described in Section \ref{sec:core} and the  performance of the different models is measured by  the MSE between the predicted mortality rates and the actual ones. 

In this setting, we define the response variable scaled in $[0,1]$ as in \cite{Perla2021}. This does not modify the general model but  induces some changes in the formulas for the NN estimates of LC parameters. Details can be found in  Appendix \ref{app:modelscaled}.

\subsubsection{Results}
\begin{figure}[htb!]
	\centering
    	\includegraphics[width=1\linewidth]{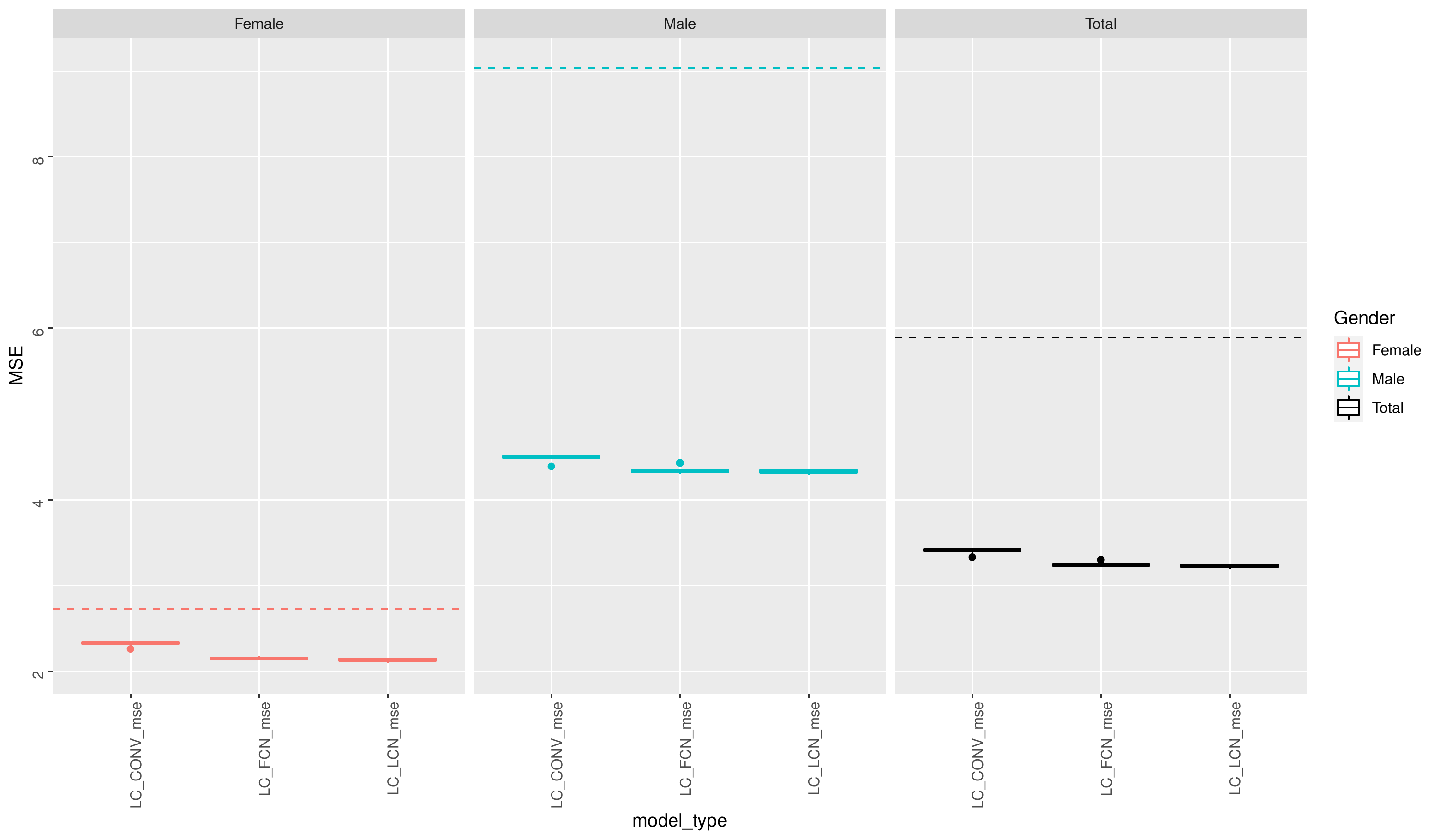}
    \caption{Box plots of the forecasting MSEs of the different models for female populations (in red), male populations (in blue) and total populations (in black); forecasting period 2000-2019; MSEs values are in $10^{-4}$. }
    \label{fig:boxplots_forecasting_mse}
\end{figure}

In this section, we  discuss the  results of the numerical experiments. 
Since the  out-of-sample  results of the neural network models can vary among training attempts due to the randomness of some elements of the training process (i.e.~the random selection of batches of training data, dropout, the initial value of optimisation algorithm and others), we first analyse and measure the variability of these results.
For this purpose, 10 different model fittings  for each  network are performed and the boxplots of the corresponding forecasting MSEs are decipted in Figure \ref{fig:boxplots_forecasting_mse}. 
 In particular,  this figure provides three subplots: the first one (with the red boxplots) concerns the MSEs obtained on female populations, the second one (with blu boxplots) refers to the male populations   while  the last subplot (with black boxplots) concerns the results of all the populations.  
We also report a dashed line in each subplot which indicates the forecasting performance  of the standard LC methodology via SVD henceforth indicated as LC\_SVD.
Some interesting comments can be made. 
First, we note that, in all the considered cases, the boxplots are  tight suggesting that the variability of the out-of-sample results of all the three neural network models is very small.
Second,  since all the boxplots lie  below the dotted line,  we conclude that  the three models produce  forecasting MSEs significantly lower than the LC\_SVD model in all training attempts. 
More in detail,  we  observe that the improvement produced by the neural network models  with respect to the  LC\_SVD model mainly concerns the male populations as the largest distance  between the boxplots and the dashed line is visible in the plot in blue. 
Finally, comparing the boxplots of the  neural network models with each other, we see that the LC\_LCN\_mse and LC\_FCN\_mse models produce very similar results while the LC\_CONV\_mse model appears to produce larger MSEs.
This evidence, which seems to work in all the cases,  probably suggests that some of the network parameters employed  by the LC\_FCN\_mse model are redundant and  the reduction in the number of parameters induced by the local connectivity does not deteriorate the performance. 
On the contrary, the further reduction induced by the weight sharing mechanism, on which the LC\_CONV\_mse model is based,  produces a slight deterioration in performance.  
We believe this may be because using the same set of parameters for all the age groups  we could reduce the quality of the  feature extraction.
This argument appears plausible since the mortality rates in different ages present different features.
Overall, we conclude that all the network models appear competitive against the LC\_SVD. Furthermore, the best result is obtained with the LC\_LCN\_mse model, which produces MSEs marginally lower than the LC\_FCN\_mse model in all three cases.  

Table \ref{tab:OLSresults} compares  the forecasting results of a single run for the network models described above. 

\begin{table}[ht]
\centering
\caption{Results of all three network architectures considered: forecasting MSE, number of populations and ages in which each network beats the LC\_SVD model; forecasting period 2000-2019; MSEs values are in $10^{-4}$. }
\medskip
\small
\begin{tabular}{| l|r|r|r|}
\hline
Model	&	\# MSE	&	\# Populations	&	\# Ages	\\
	\hline
	LC\_CONV\_mse	&	3.41	&	52	/80 &	83	/100\\
	LC\_FCN\_mse	&	3.25	&	59	/80	&	84	/100\\
	LC\_LCN\_mse	&	3.22	&	60	/80	&	84	/100\\
\hline
\end{tabular}
\label{tab:OLSresults}
\end{table}
We observe that all the network models produce  better global performance than LC\_SVD which produces an MSE equal to  $6.12 \cdot 10^{-4}$. 
Table \ref{tab:OLSresults} also lists, for each network, the number of populations and ages in which the  MSE produced is lower  than that obtained through the LC\_SVD model.
Also in this case, we observe that LC\_LCN\_mse and LC\_FCN\_mse models produce a  good performance.  
They beat the LC\_SVD model in 75\% of the populations considered and in almost  85\% of the age considered. 
Considering the forecasting performance and the number of parameters,  we select LC\_LCN\_mse model as the best one and focus on it for the next section.

In our numerical experiments we also analysed how the performance of these models changes introducing non-linearities in the  subnet which processes the time-specific data component. 
We tested the same models with  rectified linear unit (ReLU) and hyperbolic tangent (tanh) activation functions in the first layer of the third subnet. These tests, which are not presented in this paper to keep a  simple description, lead to the same results observed in \cite{Perla2021}: using the tanh activation we obtain similar results with respect to the linear one  while the ReLU activation deteriorates the performance.

\subsubsection{Estimates comparison}

\begin{sidewaysfigure}
\includegraphics[width=\columnwidth]{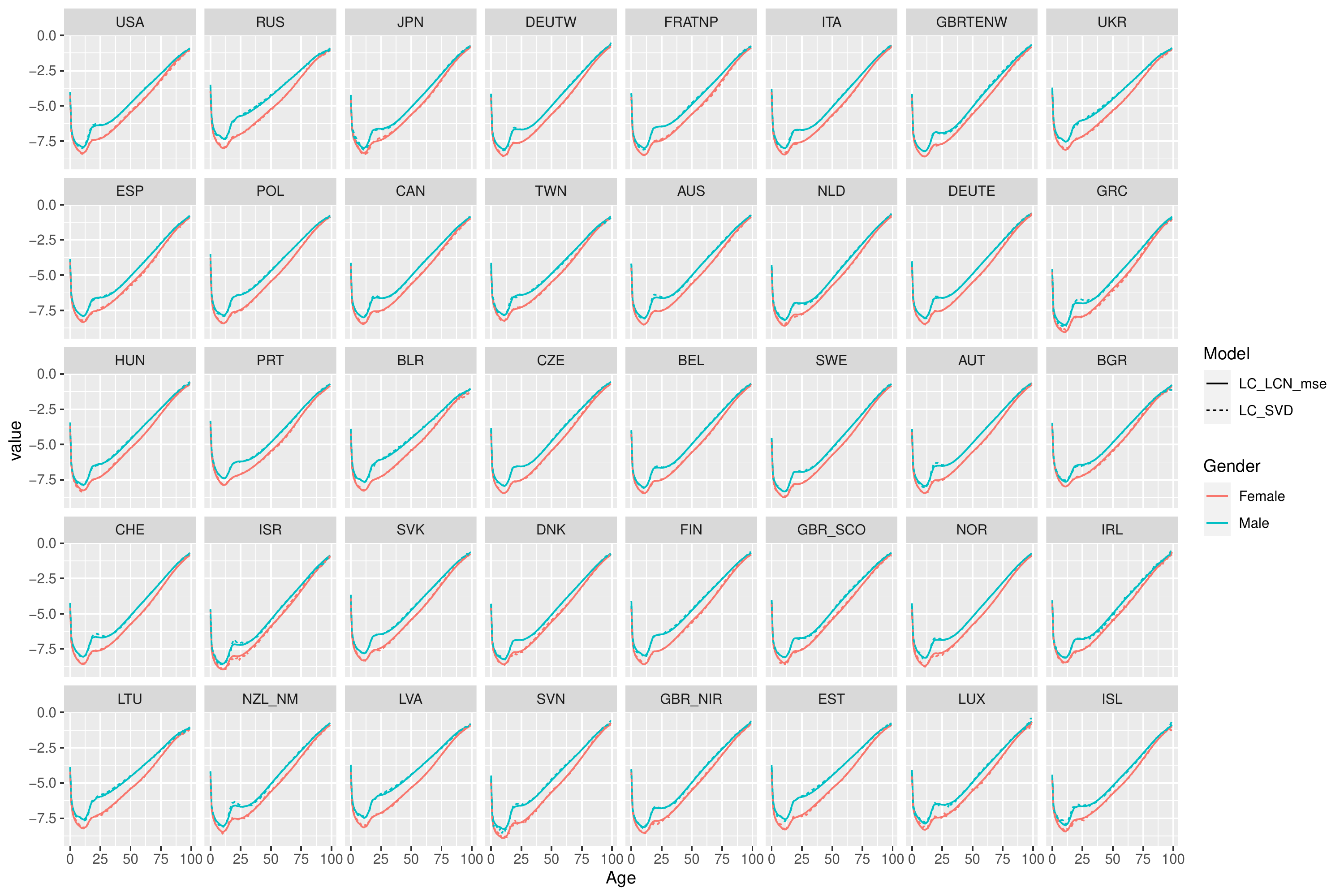}%
    \caption[angle = 90]{Comparison of the LC\_LCN\_mse and LC\_SVD estimates of $(a_x^{(i)})_{x \in \mathcal{X}}$ for all the populations considered; fitting period 1950-1999; countries are sorted by population size in 2000.}  
    \label{fig:OLS_estimator_ax}
\end{sidewaysfigure}

\begin{sidewaysfigure}
\includegraphics[width=\columnwidth]{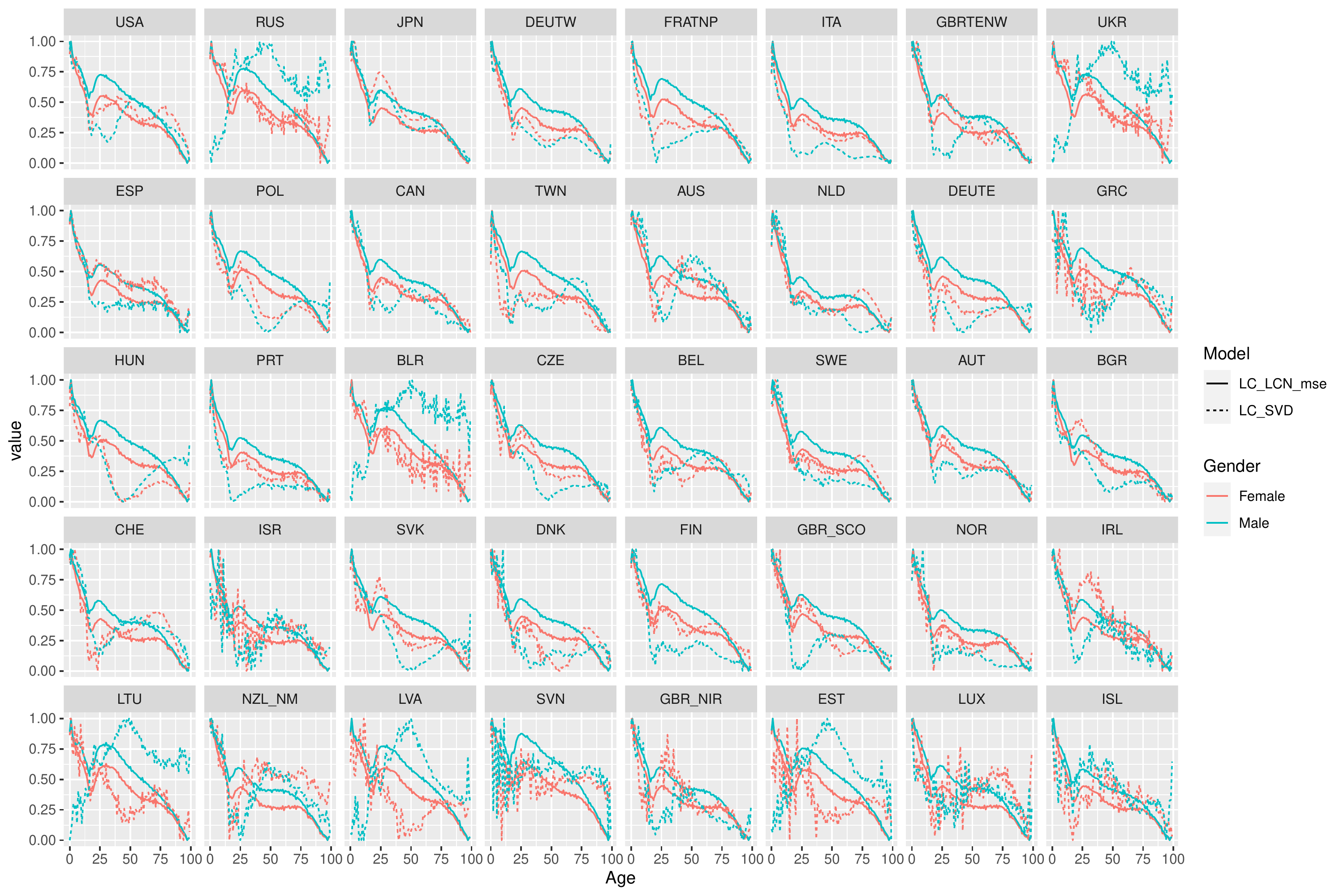}%
    \caption{Comparison of the LC\_LCN\_mse and LC\_SVD estimates of $(b_x^{(i)})_{x \in \mathcal{X}}$ for all the populations considered; fitting period 1950-1999; countries are sorted by population size in 2000.}  
    \label{fig:OLS_estimator_bx}
\end{sidewaysfigure}

\begin{sidewaysfigure}
\includegraphics[width=\columnwidth]{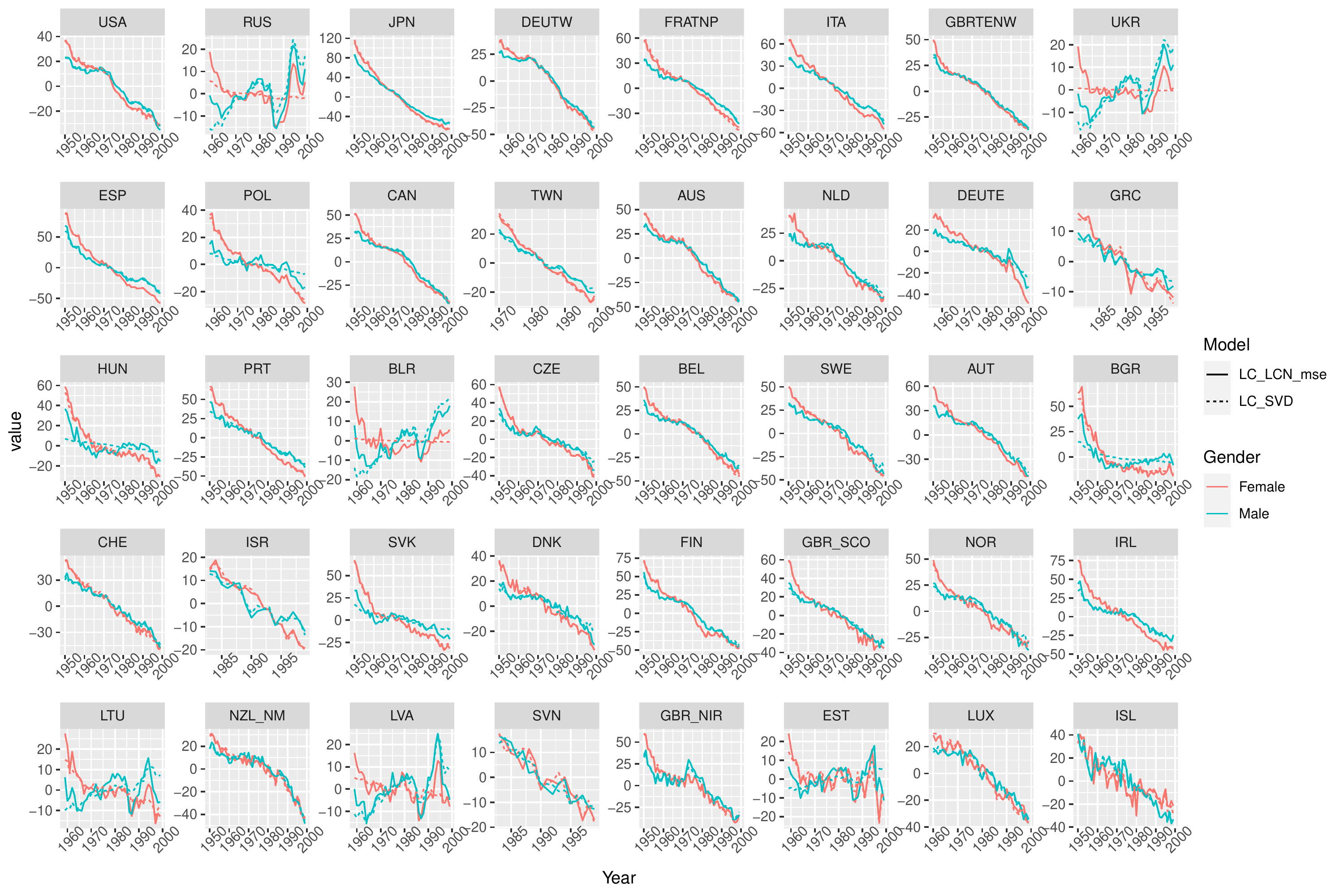}%
    \caption{Comparison of the LC\_LCN\_mse and LC\_SVD estimates of $(k_t^{(i)})_{t \in \mathcal{T}}$ for all the populations considered; fitting period 1950-1999; countries are sorted by population size in 2000.} 
    \label{fig:OLS_estimator_kx}
\end{sidewaysfigure}

In this section,  we  analyse the  estimates of the LC parameters obtained via the LC\_LCN\_mse model comparing them against the LC\_SVD approach. 
Figures \ref{fig:OLS_estimator_ax}, \ref{fig:OLS_estimator_bx} and \ref{fig:OLS_estimator_kx}  compare the  estimates of $({a}^{(i)}_x)_x$, $({b}^{(i)}_x)_x$ and $({k}^{(i)}_t)_t$   for all populations $i \in \mathcal{I}$. 
These three figures provide some country-specific subplots in which the curves of the parameters  are represented distinguishing by gender: the male population's parameters are presented in blue,  while the female population's parameters are decipted in red.
The subplots are sorted in descending order with respect to the population size in the year 2000, the first year of the testing set.
In each subplot, the solid lines refer to the LC\_LCN\_mse estimates while the dashed lines denote those obtained via LC\_SVD.
We discuss these figures one by one in the following.
 
\begin{figure}[h]
	\centering
	\includegraphics[width=1\linewidth]{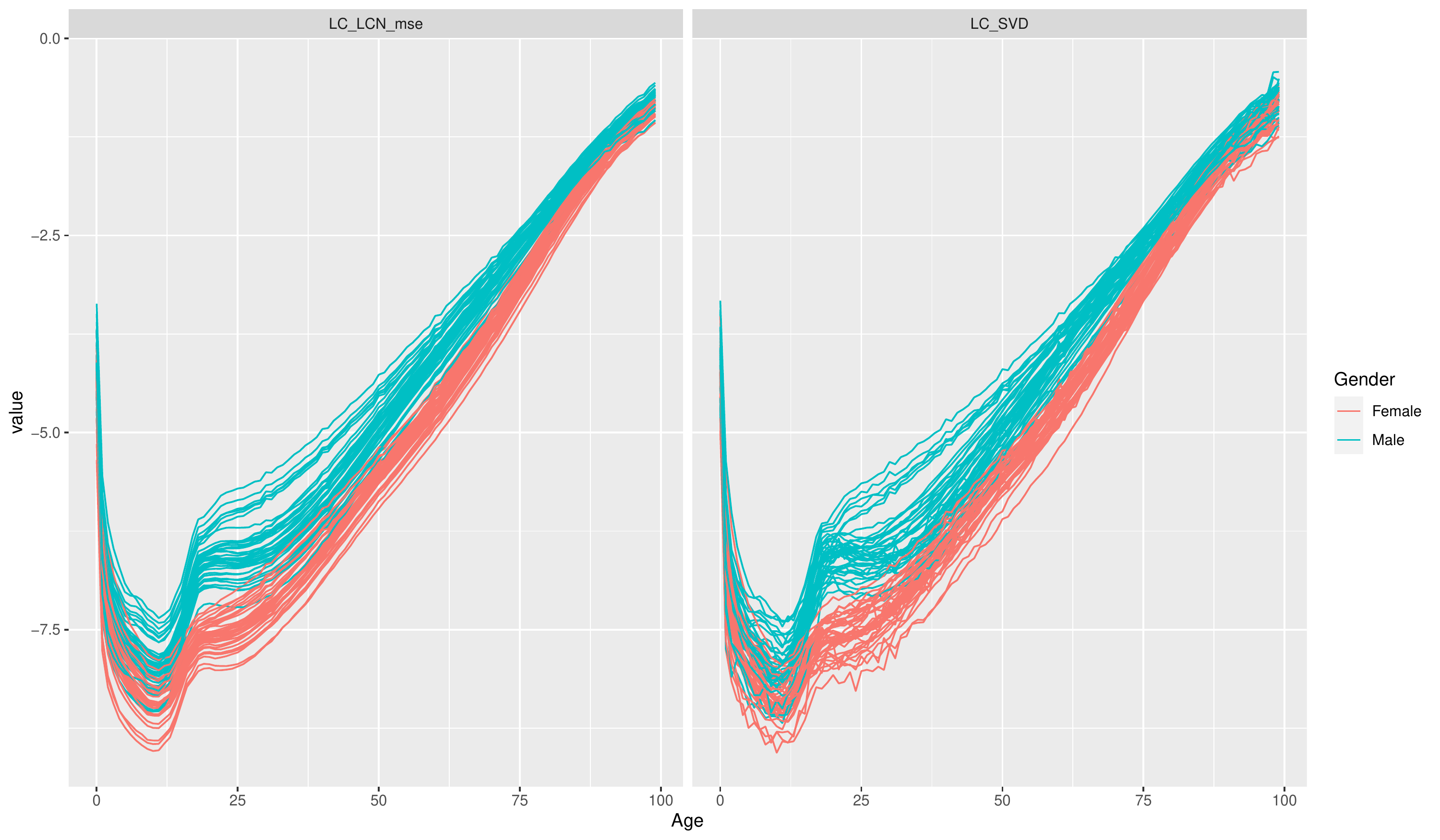}
	    \caption{Comparison of the LC\_LCN\_mse and LC\_SVD estimates of $(a_x^{(i)})_{x \in \mathcal{X}}$ distinguishing by model; fitting period 1950-1999.} 
    \label{fig:OLS_estimator_all_ax}
\end{figure}

 Figure \ref{fig:OLS_estimator_ax} compares the estimations of the $(a_x^{(i)})_x$ for all populations considered.
 First, we observe that all the  curves present the classical life-table shape and the  female  curves are located below the male ones. 
 Furthermore, since the dotted  and solid lines seem almost coincide, we  conclude that   both  approaches produce similar  $a_x^{(i)}$ estimates.
 To better understand the differences between these estimations  we analyse them more closely
by  representing the $(a_x^{(i)})_x$ curves simultaneously on the same plot. 
    Figure \ref{fig:OLS_estimator_all_ax} shows this comparison on two different subplots; the first one concerns the LC\_SVD model (right) while the second one represents the LC\_LCN\_mse model (left). 
Looking at the $(a_x^{(i)})_x$ estimations from this point of view, we observe that the LC\_SVD curves  present some erratic fluctuations while the LC\_LCN\_mse estimates appear to be smooth.
This could    due to the random fluctuations often present in the mortality data which also affect  the parameters estimates. 
We could explain this evidence by arguing that the LC\_SVD works on a population-specific subset of data, and then its estimates, could be more sensitive to  the random fluctuations present in that  data. 
On the contrary, the LC\_LCN\_mse model, which uses a large amount of data  to fit and allows the information sharing among the populations through cross-population parameters,  prevents population-specific overfitting and produces estimates  less  sensitive to these fluctuations. 
It should be remarked that the dropout, which has been used in all  three subnets, also contributes to the smoothness of the curves estimated.

Figure \ref{fig:OLS_estimator_bx} compares the  $(b_x^{(i)})_x$ estimates (scaled in  $[0,1]$).

Here, we observe that the two approaches involve quite different $(b_x^{(i)})_x$ estimations. 
This is especially evident for  low-population countries which are represented at the bottom of the figure.
In particular, we note that the LC\_SVD estimates appear to be smooth for high-population countries (e.g USA and JPN) while they  present irregular patterns for low-population  countries (e.g. LUX and ISL). 
This evidence was already discussed in the literature, see \cite{Delwarde2007}.

Also in this case, this could be due to the random fluctuations present in the mortality rates which appears more pronounced for low population countries \cite{Jarner2011}. 
We believe  this is related to the law of large numbers, which makes volatility in morality rates larger for low-population countries.
On the contrary, the  LC\_LCN\_mse curves are quite smooth for all countries.
This can be justified with the same arguments given for the $(a_x^{(i)})_x$ estimates.

Figure  \ref{fig:OLS_estimator_kx}   compares the LC\_LCN\_mse and LC\_SVD estimates of  $(k_t^{(i)})_t$ for all populations considered. In this case no particular differences emerge; the two approaches produce rather similar $(k_t^{(i)})_{t \in \mathcal{T}}$ estimates.

\subsection{Poisson Loss minimisation}

As  emphasised in Section \ref{sec:poisson_lc}, the assumption of homoskedastic error structure which follows the ordinary least squares estimation, often appears unrealistic. 
In this section, assuming  a Poisson number of death $D^{(i)}_{x,t}$, we explore the use of the Poisson loss function to train the neural network models. 
In particular, we consider 
\begin{equation}
\label{eq:poisson_assumption_NN}
D^{(i)}_{x,t}  \sim Poisson (E^{(i)}_{x,t} e^{m^{(i)}_{x,t}}),
\end{equation}
where 
\begin{equation*}
    {m}_{x,t}^{(i)} = \bigg( w^{(a)}_{x,0} +  \left\langle \bw_x^{(a)}, \bz_{\mathcal{I}}^{(a)} \right\rangle  \bigg)\ + 
    \end{equation*}
    \begin{equation}
  +  \bigg( w^{(b)}_{x,0} +\left\langle \bw_x^{(b)}, \bz_{\mathcal{I}}^{(b)} \right\rangle \bigg) \cdot \bigg( w^{(k_2)}_0 +  \left\langle \bw^{(k_2)}, \phi^{(k_1)} \bigg(  \bw^{(k_1)}_0 +  \left\langle W^{(k_1)}, log(\boldsymbol{m}_{x}^{(i)}) \right\rangle  \bigg) \right\rangle \bigg).
\end{equation}
In this setting,  the neural networks model fitting involves the  minimisation of 
\begin{equation}
L(\psi) =  \sum_{x \in \mathcal{X}}  \sum_{i \in \mathcal{I}}\sum_{t \in \mathcal{T}} \bigg(   E^{(i)}_{x,t} e^{{m}_{x,t}^{(i)}}- D^{(i)}_{x,t} {m}_{x,t}^{(i)} \bigg) + c\\
\end{equation}

which corresponds to maximise the log-likelihood function under the assumption \eqref{eq:poisson_assumption_NN} and $c \in \mathbb{R}$. We use the same data of section \ref{sec:experiments_ols}; however, this time we exclude the Canadian populations since the data present several missing values.  Here,  we have $|\mathcal{I}| = 78$. 

The   neural network models previously described are trained in the same setting: mortality experiences concerning calendar years in $\mathcal{T}_1$ are used for training,  the number of epochs was set equal to 2000 and the same training algorithm is employed. This time, the training sample contains 3498 examples since the 100 mortality experiences concerning Canadian populations are removed. The names of all the neural network models are suitably modified by replacing ``mse'' with ``Poisson''. The Keras code that defines the architecture of the network used is provided in the Appendix \ref{app:code}. 
In these experiments, we also included in the comparisons the results of the ILC approach based on the Poisson maximum likelihood estimation. We refer to the results of this approach as LC\_Poisson. 
We use again the MSE between predicted  mortality rates and the actual ones to measure the performance.   
Table \ref{tab:Poissonresults} lists the forecasting MSE  for each neural network model. The LC\_SVD and the LC\_Poisson respectively obtain MSEs equal to $6.02 \cdot 10^{-4}$ and $5.19\cdot 10^{-4}$.
\begin{table}[ht]
\centering
\caption{Results of all  three network architectures considered: forecasting MSEs, number of populations and ages in which each network beats the LC\_Poisson and the LC\_SVD models; forecasting period 2000-2019; MSEs values are in $10^{-4}$. }
\medskip
\small
\begin{tabular}{|l|r|cc|cc|}
\hline
& & \multicolumn{2}{|c|}{ LC\_Poisson} &  \multicolumn{2}{|c|}{ LC\_SVD}   \\
\hline 
Model	& MSE	&	\# Populations	&	\# Ages &	\# Populations	&	\# Ages	\\
	\hline
	LC\_CONV\_Poisson & 3.02 & 57/78 & 83/100 & 64/78 & 83/100\\
	LC\_FCN\_Poisson & 3.07 & 55/78 & 83/100 & 63/78  & 83/100\\
	LC\_LCN\_Poisson & 2.89 & 61/78 & 83/100 & 67/78 & 83/100\\
\hline
\end{tabular}
\label{tab:Poissonresults}
\end{table}
First, we observe that all three neural networks models overperform  LC\_SVD and LC\_Poisson. In particular, the LC\_LCN\_Poisson produces the best performance, LC\_CNN\_Poisson is the second one and LC\_FNC\_Poisson is the least accurate.
Table \ref{tab:Poissonresults} also reports, for each of these three models, the number of populations and ages in which it beats the LC\_SVD and LC\_Poisson models. 
Again,   we observe that all  three neural network models perform very well against the two benchmarks, especially against the LC\_SVD approach.
Overall, we select the LC\_LCN\_Poisson model  as the best model and further comparisons against LC\_Poisson are provided below. 

Table \ref{tab:comparison_by_pop} reports the  LC\_LCN\_Poisson and LC\_Poisson models' performance in all countries under investigation distinguishing by gender.
The best performance in each population is reported in bold.
We observe that, in the most of the cases, LC\_LCN\_Poisson beats the LC\_Poisson and this evidence appears especially evident for male populations. In addition, in some of these cases, the improvement produced by  LC\_LCN\_Poisson model results quite  large (see for example the male populations of LUX, SVN, SVK, BLG, UKR and others). Many of these countries are  low population countries since they are located in  the bottom of Table \ref{tab:comparison_by_pop}. 
This evidence suggests that LC\_LCN\_Poisson model could  improve the forecasting in low-population countries whose data are often affected by random fluctuations. 
On the contrary, in the cases in which the LC\_Poisson model beats the LC\_LCN\_Poisson model, the difference is often quite limited;  the only cases in which this difference appears significant are the female populations of JPN and LTU. 
The evidence shown in Table \ref{tab:comparison_by_pop} would be even more prominent if the LC\_LCN\_Poisson model was compared against the LC\_SVD model.

\begin{table}[tbp]
\centering
\caption{Forecasting MSEs of  the LC\_LCN\_Poisson, LC\_Poisson, LC\_SVD  models on different populations; forecasting period 2000-2019; MSEs values are in $10^{-4}$.}
\medskip
\small
\begin{tabular}{|ll|rr|rr|}
\hline
& & \multicolumn{2}{|c|}{Male} &  \multicolumn{2}{|c|}{Female}   \\
\hline
 & Country & LC\_LCN\_Poisson & LC\_Poisson  & LC\_LCN\_Poisson & LC\_Poisson \\
\hline
1	&	USA	&	{\bf 1.15}	&	1.42	&	{\bf 0.27}	&	0.50	\\
2	&	RUS	&	{\bf 2.19}	&	8.35	&	{\bf 2.23}	&	5.89	\\
3	&	JPN	&	0.91	&	{\bf 0.45}	&	2.30	&	{\bf 0.40}	\\
4	&	DEUTW	&	{\bf 0.63}	&	0.80	&	{\bf 0.23}	&	 0.35	\\
5	&	FRATNP	&	0.77	&	{\bf 0.52}	&	0.64	&	{\bf 0.34}	\\
6	&	ITA	&	{\bf 0.49}	&	0.58	&	0.94	&	{\bf 0.24}	\\
7	&	GBRTENW	&	{\bf 0.74}	&	1.11	&	0.66	&	{\bf 0.38}	\\
8	&	UKR	&	{\bf 2.05}	&	7.19	&	{\bf 3.40}	&	3.72	\\
9	&	ESP	&	{\bf 0.80}	&	1.72	&	{\bf 0.63}	&	1.27	\\
10	&	POL	&	{\bf 2.61}	&	4.69	&	{\bf 0.85}	&	3.29	\\
11	&	TWN	&	{\bf 4.82}	&	10.49	&	1.42	&	{\bf 0.95}	\\
12	&	AUS	&	{\bf 0.89}	&	1.14	&	{\bf 0.32}	&	0.41	\\
13	&	NLD	&	{\bf 1.11}	&	1.76	&	0.43	&	{\bf 0.35}	\\
14	&	DEUTE	&	{\bf 1.82}	&	2.71	&	{\bf 0.70}	&	1.45	\\
15	&	GRC	&	{\bf 1.73}	&	3.16	&	{\bf 0.55}	&	1.97	\\
16	&	HUN	&	{\bf 3.45}	&	6.01	&	{\bf 1.22}	&	1.38	\\
17	&	PRT	&	{\bf 1.33}	&	2.42	&	{\bf 0.99}	&	2.01	\\
18	&	BLR	&	{\bf 3.34}	&	12.76	&	{\bf 3.47}	&	10.24	\\
19	&	CZE	&	{\bf 2.97}	&	4.68	&	{\bf 1.03}	&	2.27	\\
20	&	BEL	&	{\bf 1.56}	&	2.31	&	{\bf 0.47}	&	0.51	\\
21	&	SWE	&	{\bf 1.10}	&	1.13	&	{\bf 0.25}	&	0.38	\\
22	&	AUT	&	{\bf 1.51}	&	2.57	&	{\bf 0.40}	&	0.61	\\
23	&	BGR	&	{\bf 5.83}	&	11.30	&	{\bf 2.95}	&	6.14	\\
24	&	CHE	&	{\bf 1.41}	&	1.81	&	0.32	&	{\bf 0.32}	\\
25	&	ISR	&	2.38	&	{\bf 1.85}	&	2.03	&	{\bf 1.81}	\\
26	&	SVK	&	{\bf 7.20}	&	13.27	&	3.20	&	{\bf 2.54}	\\
27	&	DNK	&	{\bf 2.01}	&	2.27	&	0.53	&	{\bf 0.42}	\\
28	&	FIN	&	3.74	&	{\bf 3.73}	&	{\bf 0.82}	&	1.10	\\
29	&	GBR\_SCO	&	{\bf 1.69}	&	1.97	&	{\bf 0.41}	&	0.67	\\
30	&	NOR	&	{\bf 2.11}	&	3.50	&	0.71	&	{\bf 0.51}	\\
31	&	IRL	&	{\bf 3.40}	&	7.82	&	{\bf 1.52}	&	2.23	\\
32	&	LTU	&	{\bf 6.59}	&	9.37	&	9.54	&	{\bf 7.60}	\\
33	&	NZL\_NM	&	{\bf 2.50}	&	4.19	&	{\bf 0.70}	&	1.19	\\
34	&	LVA	&	{\bf 10.38}	&	11.37	&	{\bf 3.00}	&	3.57	\\
35	&	SVN	&	{\bf 10.18}	&	69.32	&	{\bf 2.01}	&	4.77	\\
36	&	GBR\_NIR	&	{\bf 5.75}	&	8.21	&	{\bf 1.62}	&	1.80	\\
37	&	EST	&	{\bf 16.05}	&	18.88	&	{\bf 3.49}	&	6.88	\\
38	&	LUX	&	{\bf 15.90}	&	43.12	&	{\bf 5.42}	&	6.74	\\
39	&	ISL	&	{\bf 19.17}	&	19.98	&	7.56	&	{\bf 7.40}	\\

\hline
\end{tabular}
\label{tab:comparison_by_pop}
\end{table}

\begin{figure}[htb!]
	\centering
	\includegraphics[width=1\linewidth]{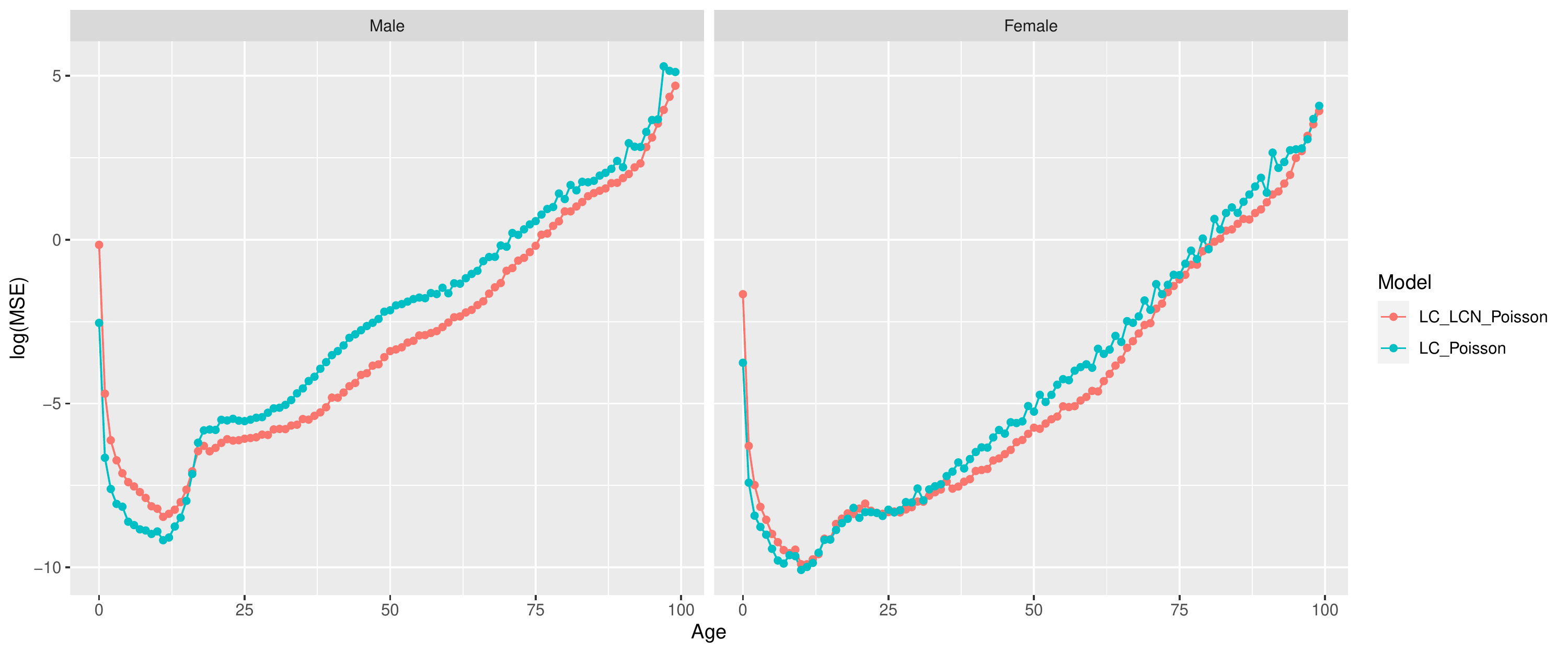}
    \caption{Comparison error between the LC\_LCN\_Poisson and the LC\_Poisson model for years in 2000-2019 for different ages and genders, MSE values are in $10^{-4}$ and on log-scale.}
    \label{fig:Poisson_error_by_age}
\end{figure}

Figure \ref{fig:Poisson_error_by_age} compares the LC\_LCN\_Poisson and LC\_Poisson models by analysing the error by age for both genders on a logarithmic scale. In both cases, the curves show the shape of a mortality table. Furthermore,  we observe that, for both genders, the LC\_LCN\_Poisson model produces a lower forecasting error for  ages $x>20$ and  this improvement is more evident for males. On the contrary, for ages $x\leq 20$ the LC\_Poisson beats the LC\_LCN\_Poisson model. 
We conclude that the LC\_LCN\_Poisson model presents an overall forecasting performance higher than the LC\_Poisson and LC\_SVD models.

\subsection{Some sensitivity tests}
In this section, the LC\_LCN\_Poisson model is submitted to some sensitivity checks. First, we analyse the sensitivity of the LC\_LCN\_Poisson model respect to the set of populations considered $\mathcal{I}$. The forecasting performance should not  roughly change  when a smaller set of populations is considered.
To this aim, we split the full set of populations $\mathcal{I}$ into two subsets, the first one, $\mathcal{I}_{HP}$, contains the male and female populations the 20 highest population countries (from 1-20 of Table \ref{tab:comparison_by_pop}) while the second set,  $\mathcal{I}_{LP}$,  includes the male and female populations of the remaining 19 the lowest population countries such that $\mathcal{I} = \mathcal{I}_{LP} \cup \mathcal{I}_{HP}$. 
\begin{table}[ht]
\centering
\caption{Forecasting MSEs of  the LC\_LCN\_Poisson, LC\_Poisson, LC\_SVD  models on $\mathcal{I}_{HP}$ and $\mathcal{I}_{LP}$; forecasting period 2000-2019; MSEs values are in $10^{-4}$.}
\medskip
\small
\begin{tabular}{|l|rrr|}
\hline
 & LC\_LCN\_Poisson & LC\_Poisson &  LC\_SVD \\
	\hline
	$\mathcal{I}_{LP}$	&	4.49	&	7.46 &	8.97\\
	$\mathcal{I}_{HP}$	&	1.41	& 2.74	 & 2.86 \\
\hline
\end{tabular}
\label{tab:countries_split}
\end{table}

The model LC\_LCN\_Poisson model is run separately on these two sets and the results have been collected in Table \ref{tab:countries_split}. 
Again, the results for the LC\_SVD and LC\_Poisson models are reported to make comparisons. 
We observe that on the high-population countries  all the models produce  lower MSEs than that obtained on the full set $\mathcal{I}$.
On the contrary, on the low-population countries we observe that the respective MSEs are significantly higher. In both cases we note that the LC\_LCN\_Poisson model still overperforms the two benchmarks.
Therefore, we  conclude that the LC\_LCN\_Poisson model is definitely competitive even when smaller sets of populations are considered.
Nevertheless, it is reasonable to believe that the LC\_LCN\_Poisson model benefits from using as much data as possible for the training. 

\begin{figure}[tbp]
	\centering
	\includegraphics[width=1\linewidth]{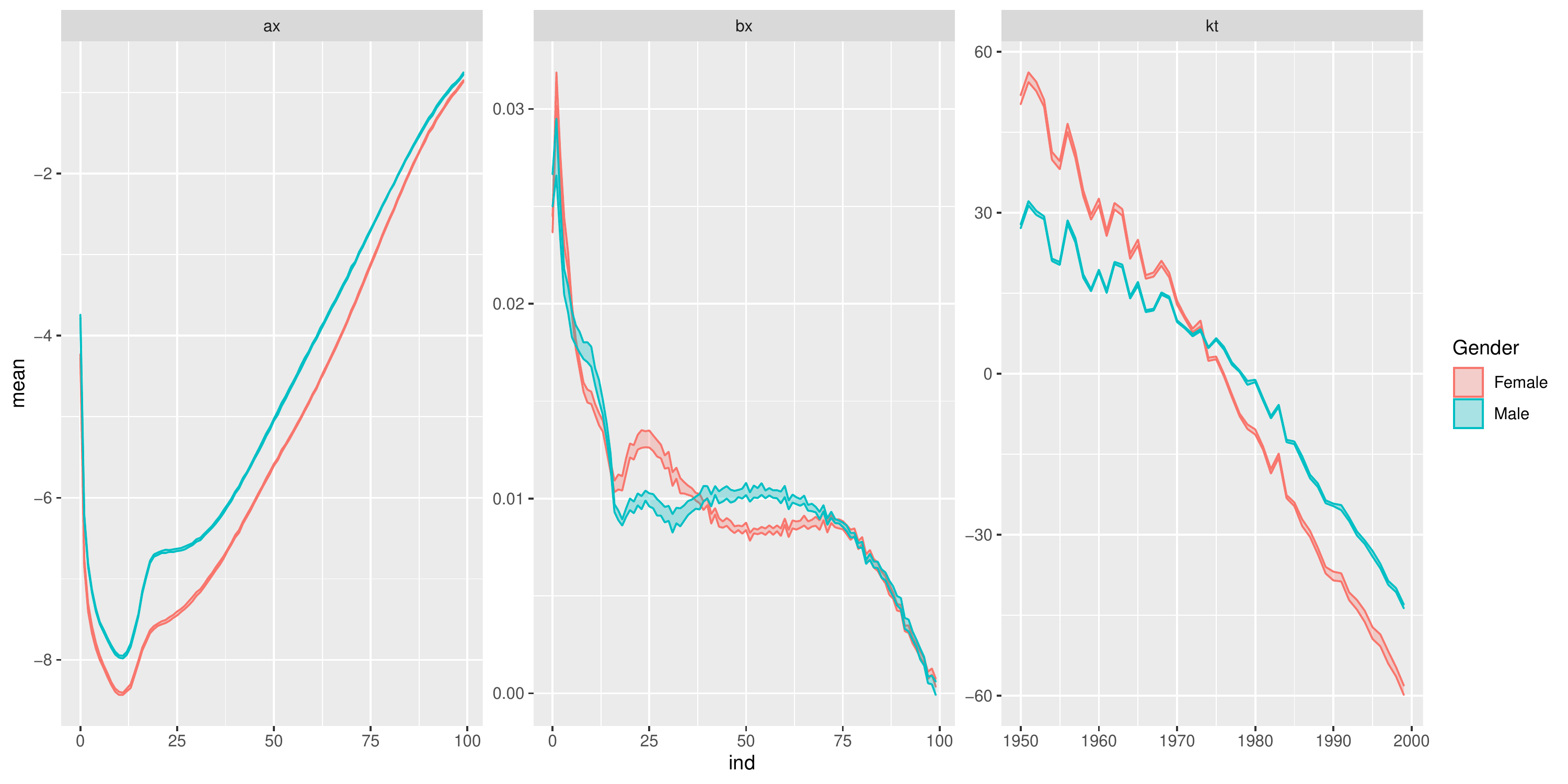}
	    \caption{Variability in the  estimates of $(a_x^{(i)})_x$, $(b_x^{(i)})_x$ and $(k_t^{(i)})_t$  obtained by the LC\_LCN\_Poisson model in 10 training attempts  for Italian populations; fitting period 1950-1999. }
    \label{fig:Poisson_parameters_ITA}
\end{figure}

In the second part of this section, we investigate how much  the LC\_LCN\_Poisson model's results  change in the different training attempts. 
 As highlighted in  Section \ref{sec:NN}, the results of training a neural network are somewhat variable, and, therefore, the estimation of the optimal network parameters $\hat{\psi}$ can vary  between training attempts.
This affects also the  NN estimations of the LC parameters which  vary themselves between  training attempts. 
In this section we investigate the variability of these estimates keeping in mind that  a small variability means that the results produced by the neural network model are stable.
On the contrary, a   large  variability could highlight that the results produced by the network change significantly between the different training attempts and then they are not stable. 
In order to investigate this point, we consider the NN estimations of the LC parameters   obtained in the 10 different training attempts of the LC\_LCN\_Poisson model. 
Figure \ref{fig:Poisson_parameters_ITA} represents the full ranges of variation of the NN estimations of the LC parameters for the Italian populations. It includes three subplots which refer to the $(a_x^{(i)})_x$, $(b_x^{(i)})_x$ and $(k_t^{(i)})_t$  parameters respectively.
Two lines are decipted for each parameter curve; the higher curve represents the maximum values observed in the different training attempts while the lower one represents the minimum values. Intuitively, the area  between these two bounds is the range of observed values in the 10 different runs. The graphical elements in blue refer to the male population's parameters while those in red refer to the female population's ones. 
Overall, the estimates of the LC parameters obtained seem to vary not so much. 
First, we observe that the maximum and minimum values obtained for the parameters $a_x$ and $k_t$ for the Italian populations are  almost overlapped  and this means that the estimates obtained in the 10 training attempts are almost identical. 
About the $(b_x^{(i)})_{x}$ parameters, we note that the variability appears marginally greater than the other two parameters  for the middle ages. 
Nonetheless, since the area between the two curves is very limited also in these cases,  we can conclude  that the estimates obtained via LC\_LCN\_Poisson model are stable. 
It should be remarked that these graphs only take into account the uncertainty due to network training. 

In Appendix \ref{app:variability_plots}, Figures \ref{fig:Poisson_estimator_ax}, \ref{fig:Poisson_estimator_bx}, \ref{fig:Poisson_estimator_kt}  extend this analysis to all the other populations. Figure  \ref{fig:Poisson_estimator_ax} depicts the ranges of the  $(a^{(i)}_x)_x$ estimates for the different populations; Figures
\ref{fig:Poisson_estimator_bx} and \ref{fig:Poisson_estimator_kt} analyse respectively the  estimates $(b^{(i)}_x)_x$ and $(k^{(i)}_t)_t$. 

The findings  obtained for the Italian populations appear to work for  all other countries. The estimates obtained for the LC parameters $a^{(i)}_x$ and $k_t^{(i)}$ results are not very variable for all the populations considered. 
The estimates of the  $b^{(i)}_x$ result  more variable than the other parameters, especially for low-population countries. However, as shown in Figure \ref{fig:Poisson_estimator_bx}, the variability increases only for some ages and  in a marginal way.

\section{Conclusions}
\label{sec:conclusions}
This paper proposes a neural network approach for calibrating  the ILC models of multiple populations. 
The parameters of the ILC models are jointly estimated through a neural network that simultaneously processes the mortality data of all populations. 
In this way,  each individual LC model is calibrated  by exploiting all the available information instead of using a population-specific subset of data as in the traditional fitting approaches.
We experiment our approach on the HMD data considering different network architectures and loss functions, analysing the reasonableness of the parameters estimates  and the resulting forecasting performance. 
From a forecasting perspective, the numerical results show that all the network models considered overperform  the  traditional LC\_SVD and LC\_Poisson approaches for a large set of populations.
The best performance is obtained from the LC\_LCN\_Poisson model which employs locally-connected layers to extract features from the mortality rate curves. 
In particular, the forecasting performance of the LC\_LCN\_Poisson model (around $2.94 \cdot  10^{-4}$)  results comparable to the DEEP5 model (around $3.05 \cdot  10^{-4}$) proposed in \cite{Richman2020}  and  marginally poorer than  the LCCONV model (around $2.39\cdot 10^{-4}$) proposed \cite{Perla2021}. 
The LC\_LCN\_Poisson model, in addition is very efficient from the number of parameters perspective ($2.801$ parameters against $73.676$ of the DEEP5 model and $27.120$ of the LCCONV model), presents two important advantages. 
First, as discussed in Section \ref{sec:core},  it is easy-to-understand as  the network components can be interpreted. 
Second, the LC\_LCN\_Poisson model does not modify the time-series part of the LC model and it is therefore possible to derive interval estimates for the forecast mortality rates. 
Numerical experiments also show that, differently from the traditional fitting schemes, our approach produces smoother estimates of the age-specific LC  parameters curves. 
This result appears evident for the low-populations countries in which the random fluctuations in mortality rates affect the LC\_SVD and the LC\_Poisson estimates.
This could  also be the case of annuity portfolios or pension funds' data often collected considering small populations \cite{Hunt2017}. 
Future research will proceed in different directions. 
First we intend to analyse the performance of the proposed model on other available data sources such as the United State Mortality Database (USMB) and insurance portfolio's data. Second, we will investigate the use of neural networks for fitting other  stochastic mortality models. Both the single-population models belonging to the family of Generalized Age Period Cohort (GAPC) models \cite{Villegas2018} and the multi-population extensions of the LC model such as Li-Lee \cite{Li2005} and Kleinow \cite{Kleinow2015} models could be considered. 
Finally, we intend  to explore  the potential of the proposed large-scale mortality model in the actuarial evaluations and  longevity risk management.

\appendix
\section{Data Details}
\label{app:data_description}
\begin{table}[h!]
\centering
\medskip
\small
\begin{tabular}{|rr|ll||rr|ll|}
\hline
	& Country & Starting year &	Final year&	& Country & Starting year &	Final year\\

	\hline\hline
1	&	AUS	&	1950	&	2018	&	21	&	IRL	&	1950	&	2017	\\
2	&	AUT	&	1950	&	2017	&	22	&	ISL	&	1950	&	2018	\\
3	&	BEL	&	1950	&	2018	&	23	&	ISR	&	1983	&	2016	\\
4	&	BGR	&	1950	&	2017	&	24	&	ITA	&	1950	&	2017	\\
5	&	BLR	&	1959	&	2018	&	25	&	JPN	&	1950	&	2019	\\
6	&	CAN	&	1950	&	2018	&	26	&	LTU	&	1959	&	2019	\\
7	&	CHE	&	1950	&	2018	&	27	&	LUX	&	1960	&	2019	\\
8	&	CZE	&	1950	&	2018	&	28	&	LVA	&	1959	&	2017	\\
9	&	DEUTE	&	1956	&	2017	&	29	&	NLD	&	1950	&	2018	\\
10	&	DEUTW	&	1956	&	2017	&	30	&	NOR	&	1950	&	2018	\\
11	&	DNK	&	1950	&	2019	&	31	&	NZL\_NM	&	1950	&	2008	\\
12	&	ESP	&	1950	&	2018	&	32	&	POL	&	1958	&	2018	\\
13	&	EST	&	1959	&	2017	&	33	&	PRT	&	1950	&	2018	\\
14	&	FIN	&	1950	&	2019	&	34	&	RUS	&	1959	&	2014	\\
15	&	FRATNP	&	1950	&	2018	&	35	&	SVK	&	1950	&	2017	\\
16	&	GBRTENW	&	1950	&	2018	&	36	&	SVN	&	1983	&	2017	\\
17	&	GBR\_NIR	&	1950	&	2018	&	37	&	SWE	&	1950	&	2019	\\
18	&	GBR\_SCO	&	1950	&	2018	&	38	&	TWN	&	1970	&	2019	\\
19	&	GRC	&	1981	&	2017	&	39	&	UKR	&	1959	&	2013	\\
20	&	HUN	&	1950	&	2017	&	40	&	USA	&	1950	&	2018	\\

\hline
\end{tabular}
\caption{List of selected countries in $\mathcal{R}$ with the respectively initial and final years considered.}
\label{tab:countries_in_R}
\end{table}

\section{Model specification for response variable in $[0,1]$}
\label{app:modelscaled}
Sometimes, machine learning models are developed defining the  response variable scaled in $[0,1]$ by applying the MinMax scaling.
In this case, the model in \eqref{eq:full_model_compressed_matrix} can be rewritten as 
\begin{equation}
\begin{split}
\frac{\widehat{ log({m}_{x,t}^{(i)})}- y_m}{y_M - y_m}  =  \bff^{(a)} \big(  \bz_{\mathcal{I}}^{(a)} \big)  +   \bff^{(b)} \big(  \bz_{\mathcal{I}}^{(b)} \big)   
   (f^{(k_2)} \circ \bff^{(k_1)}) (log(\boldsymbol{m}_{t}^{(i)})),
\end{split}
\end{equation}
where
\[ y_m =  \min_{{x \in \mathcal{X}, t \in \mathcal{T}, i \in \mathcal{I}}} log(m_{x,t}^{(i)}) \quad \quad y_M =  \max_{{x \in \mathcal{X}, t \in \mathcal{T}, i \in \mathcal{I}}} log(m_{x,t}^{(i)}).  \]

The network parameters are calibrated minimising 
\begin{equation*}
L(\psi) =  \sum_{x \in \mathcal{X}} \sum_{i \in \mathcal{I}}  \sum_{t \in \mathcal{T}} \bigg(  \frac{{ log(m_{x,t}^{(i)})}- y_m}{y_M - y_m}  - \phi^{(a)}\bigg( w^{(a)}_{x,0} +  \left\langle \bw_x^{(a)}, \bz_{\mathcal{I}}^{(a)} \right\rangle  \bigg) +
\end{equation*}
\begin{equation}
 - \phi^{(b)}\bigg( w^{(b)}_{x,0} +  \left\langle \bw_x^{(b)}, \bz_{\mathcal{I}}^{(b)} \right\rangle \bigg) \cdot 
 \phi^{(k_2)} \bigg( w^{(k_2)}_0 +  \left\langle \bw^{(k_2)}, \phi^{(k_1)} \bigg(  \bw^{(k_1)}_0 +  \left\langle W^{(k_1)}, log(\boldsymbol{m}_{x}^{(i)}) \right\rangle  \bigg) \right\rangle \bigg)\bigg)^2.  
\end{equation}
and the correspoding NN estimates of the LC parameters in the original scale can be computed as
\begin{equation}
    \hat{a}_{x, NN}^{(i)} =  \phi^{(a)} \bigg( \hat{w}_{x, 0}^{(a)}+\left\langle \hat{\bw}_{x,\mathcal{R}}^{(a)}, \hat{\bz}_{\mathcal{R}}^{(a)} (r) \right\rangle + \left\langle \hat{\bw}_{x,\mathcal{G}}^{(a)} , \hat{\bz}_{\mathcal{G}}^{(a)} (g) \right\rangle \bigg) \bigg(y_M - y_m\bigg) + y_m,
\end{equation}
\begin{equation}
    \hat{b}_{x, NN}^{(i)} = \phi^{(b)} \bigg( \hat{w}_{x, 0}^{(b)}+\left\langle \hat{\bw}_{x,\mathcal{R}}^{(b)},  \hat{\bz}_{\mathcal{R}}^{(b)} (r) \right\rangle + \left\langle \hat{\bw}_{x,\mathcal{G}}^{(b)},  \hat{\bz}_{\mathcal{G}}^{(b)} (g) \right\rangle \bigg), 
\end{equation}
\begin{equation}
    \hat{k}_{t, NN}^{(i)} =   \phi^{(k_2)} \bigg( \hat{w}^{(k_2)}_0 +  \left\langle \hat{\bw}^{(k_2)}, \phi^{(k_1)} \bigg(  \hat{\bw}^{(k_1)}_0 +  \left\langle \hat{W}^{(k_1)}, log(\boldsymbol{m}_{x}^{(i)}) \right\rangle  \bigg) \right\rangle \bigg) \bigg(y_M - y_m\bigg). 
\end{equation}

\section{Keras Code of the LC\_LCN\_Poisson Model}
\label{app:code}

\section{Robustness of the results}
\label{app:variability_plots}

 \begin{sidewaysfigure}
\includegraphics[width=\columnwidth]{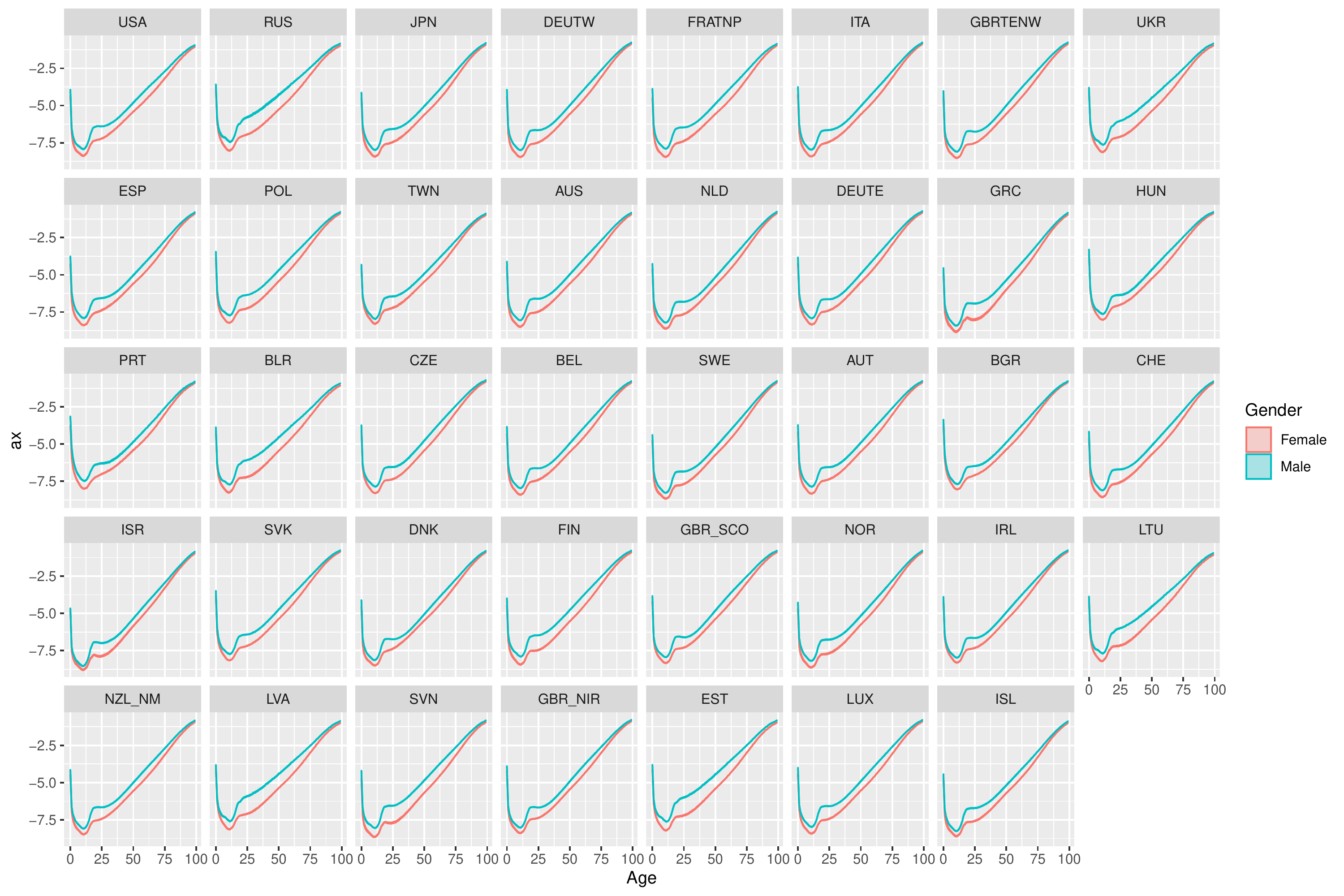}%
    \caption{Variability in the $(a_x^{(i)})_x$ estimates  obtained by the LC\_LCN\_Poisson model in 10 training attempts  for all the populations considered; fitting period 1950-1999; countries are sorted by population size in 2000. }
    \label{fig:Poisson_estimator_ax}
\end{sidewaysfigure}

 \begin{sidewaysfigure}
\includegraphics[width=\columnwidth]{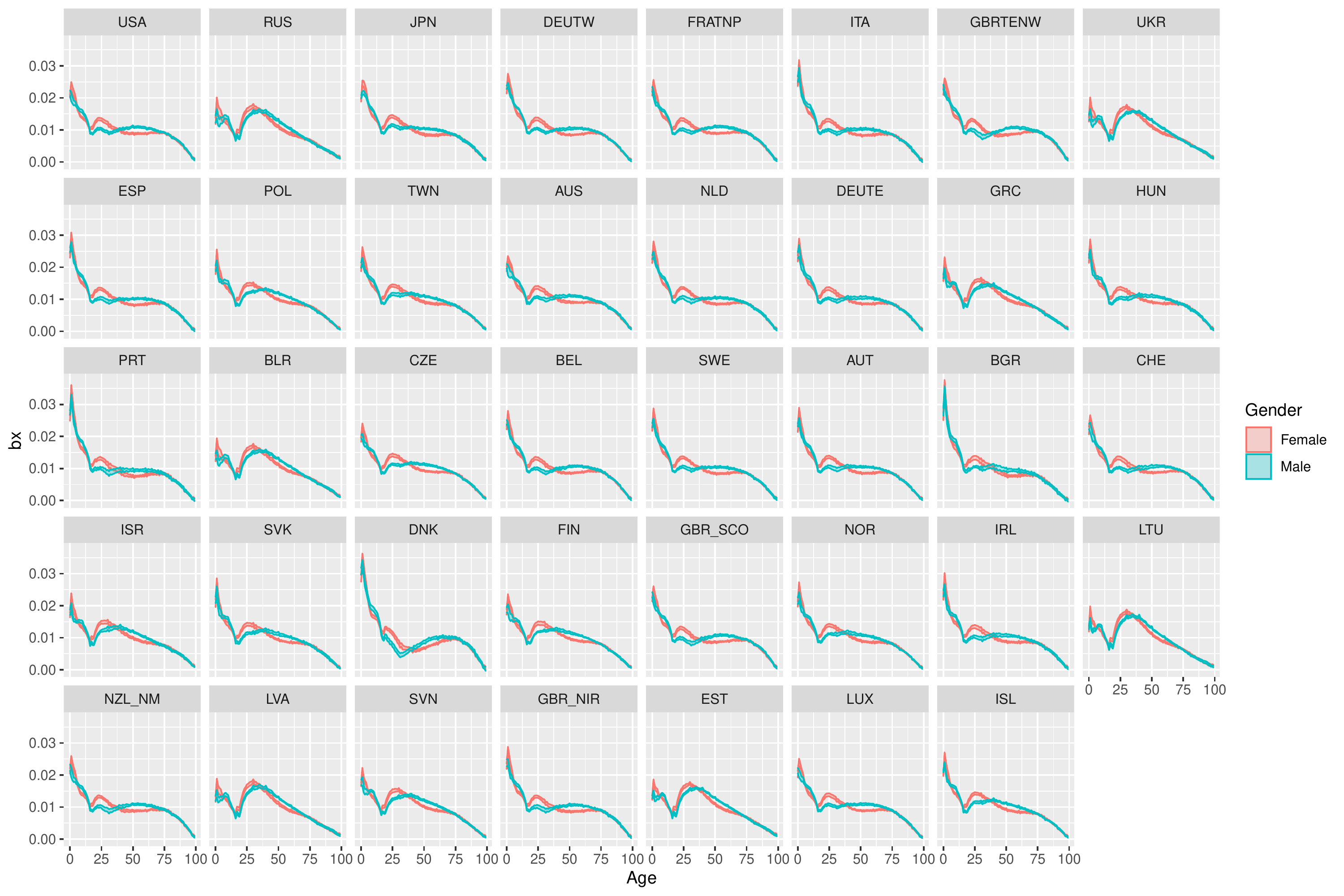}%
     \caption{Variability in the $(b_x^{(i)})_x$ estimates  obtained by the LC\_LCN\_Poisson model in 10 training attempts  for all the populations considered; fitting period 1950-1999; countries are sorted by population size in 2000. }
    \label{fig:Poisson_estimator_bx}
\end{sidewaysfigure}

 \begin{sidewaysfigure}
\includegraphics[width=\columnwidth]{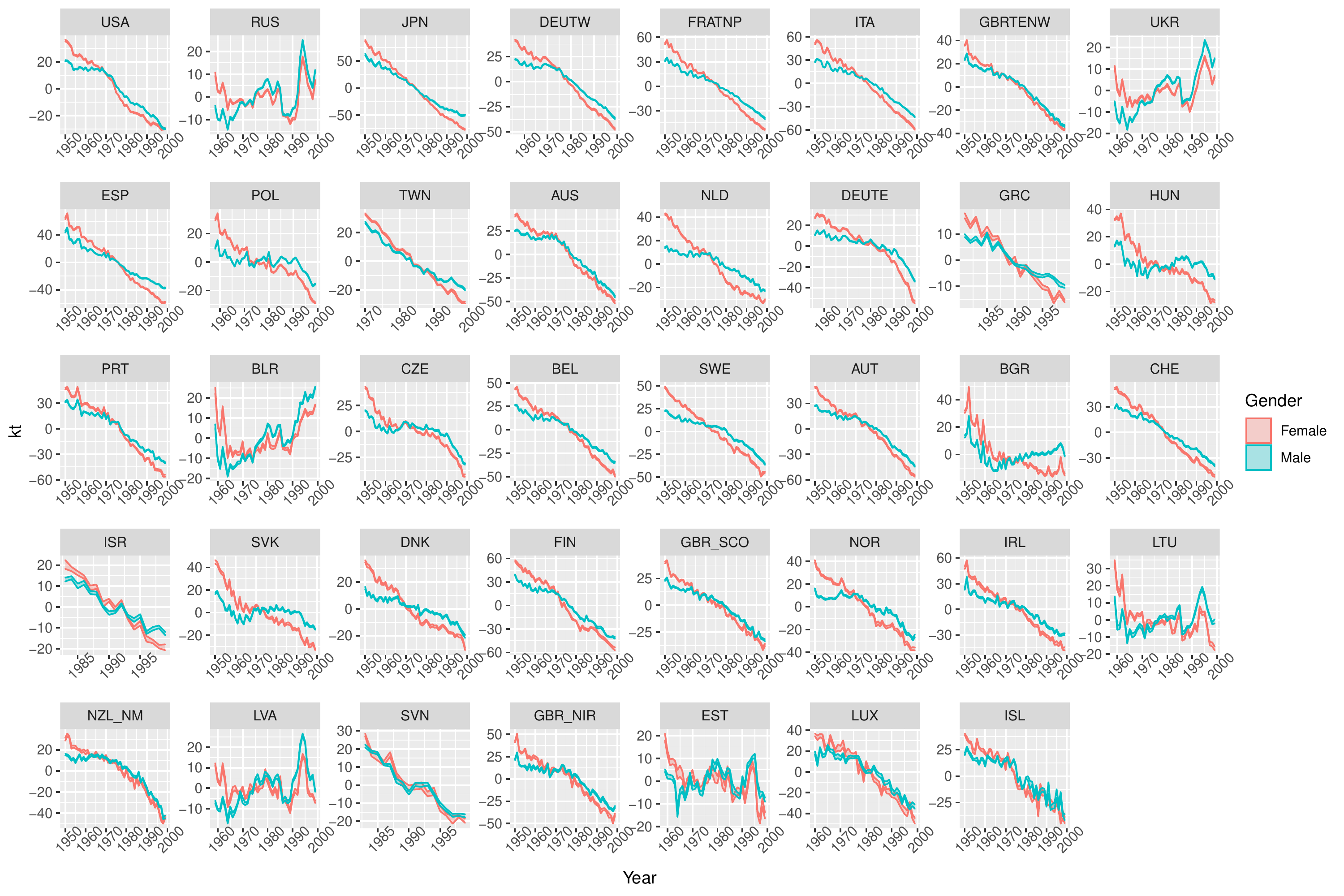}%
    \caption{Variability in the $(k_t^{(i)})_t$ estimates  obtained by the LC\_LCN\_Poisson model in 10 training attempts  for all the populations considered; fitting period 1950-1999; countries are sorted by population size in 2000. }
    \label{fig:Poisson_estimator_kt}
\end{sidewaysfigure}

 \bibliographystyle{elsarticle-num} 


{\small
	\renewcommand{\baselinestretch}{.51}
	}
\end{document}